\pdfoutput=1

\documentclass[11pt]{article}

\usepackage[preprint]{acl}

\usepackage{times}
\usepackage{latexsym}

\usepackage{graphicx}
\usepackage{bm}
\usepackage{amsmath}
\usepackage{amssymb}
\usepackage{multirow}
\usepackage{colortbl}
\usepackage{booktabs}
\usepackage{longtable}
\usepackage{adjustbox}
\usepackage{makecell}

\usepackage[T1]{fontenc}

\usepackage[utf8]{inputenc}

\usepackage{microtype}

\usepackage{inconsolata}

\title{On the Tip of the Tongue: Analyzing Conceptual Representation in Large Language Models with Reverse-Dictionary Probe}

\author{Ningyu Xu$^{1,2}$, Qi Zhang$^{1}$, Menghan Zhang$^{2,3}$, Peng Qian$^{4}$\thanks{These authors contribute equally.}, Xuanjing Huang$^{1,2}$\footnotemark[1]\\
$^{1}$School of Computer Science, Fudan University\\
$^{2}$Institute of Modern Languages and Linguistics, Fudan University\\
$^{3}$Research Institute of Intelligent Complex Systems, Fudan University\\
$^{4}$Department of Psychology, Harvard University\\
  \texttt{nyxu22@m.fudan.edu.cn} \hspace{0.01\textwidth}
  \texttt{\{qz,mhzhang,xjhuang\}@fudan.edu.cn} \hspace{0.01\textwidth} \texttt{pqian@fas.harvard.edu}
}

\begin{document}
\maketitle
\begin{abstract}

Probing and enhancing large language models' (LLMs') reasoning capacity remains a crucial open question. Here we re-purpose the reverse dictionary task as a case study to probe their capacity for conceptual inference. We use in-context learning to guide the models to generate the term for an object concept implied in a linguistic description. Models robustly achieve high accuracy in this task, and their representation space encodes information about object categories and fine-grained features. Further experiments suggest that the conceptual inference ability as probed by the reverse-dictionary task predicts model's general reasoning performance across multiple benchmarks, despite similar syntactic generalization behaviors across models. Explorative analyses suggest that prompting LLMs with description$\Rightarrow$word examples may induce generalization beyond surface-level differences in task construals and facilitate models on broader commonsense reasoning problems.

\end{abstract}

\section{Introduction}

Imagine your friend was telling a story about  their hiking trip: ``\textit{I glimpsed some sharp spikes before it quickly disappeared into the woods.}'' What was your friend talking about? You probably felt quite certain that it was not a sea urchin. But was it a hedgehog, or a porcupine, you might be wondering. Perhaps you decided to ask a question: ``\textit{How long were these spikes?}''.

As common and intuitive as the opening example, our everyday language use builds on the concepts in the mind. People's exchange of words are not merely associative responses: 
Through the chosen description of aspects of the intended referent such as ``\textit{sharp spikes}'' and ``\textit{into the woods}'', the speaker informs the listener about an object that was absent from the immediately perceived context. By building mental representations of the possibly intended referent from minimally what others say, a listener can then articulate the intended referent, form relevant questions to seek more information, and further reason about and interact with the world through words.

While concepts are ``the glue that holds our mental world together'' \citep{murphy2004big}, it remains an open question whether human-like conceptual representations and reasoning capacities emerge from statistical learning on linguistic input alone. Specifically, the contemporary large language models (LLMs) appear to be highly performant on various language comprehension and reasoning tasks after trained on gigantic amount of texts with the main objective of predicting the next token \citep{bubeck2023sparks, wei2022emergent, webb_emergent_2023, hagendorff_human-like_2023, han2024inductive}. A fruitful line of works has investigated the large language models' representation of words of specific domains, such as color \citep{patel2022mapping}, space and time \citep{gurnee2023language,geiger2023finding}, and world states in a game \citep{li2023emergent}. These works revealed impressive structural similarities between the conceptual space that a model formed contextually and its analog in the physical world where these concepts are grounded. Other works have developed synthetic tasks and datasets to evaluate the extent to which the model representations fulfill critical aspects of concepts in the human mind, such as systematic compositionality \citep{lovering-pavlick-2022-unit}. Despite the continuing efforts and progress in probing large language model's internal representation, it has been challenging to connect the model's capacity of constructing conceptual space for certain domains to a more general problem of conceptual inference, where the underlying concepts are not stated explicitly but has to be inferred from the context.

Here we develop a case study that evaluates large language models' capacity for conceptual inference and explores potential implications of such capacity on model's generalization behaviors. Inspired by the everyday referential use of language---as demonstrated in the opening example---we re-purpose the classic reverse-dictionary task and existing datasets of lexical semantics as a probe for conceptual representation in large language models. We consider the reverse-dictionary task as a special case and convenient instantiation of a general probabilistic inference problem: retrieving a lexical entry for the underlying concept given the information in a linguistic description, such as producing the word ``\texttt{dog}'' in virtue of inferring the underlying concept \textsc{dog} given the description ``\texttt{A domesticated descendant of the wolf.}'' This task itself is simple yet ecologically relevant to human communication. Consider a writer who strategically creates suspense in a story, or a person who uses words to paint an image of an object in their mind after struggling to find the exact word or phrase that names the object. Unlike previous studies where language models output meaning representation given a particular word, this word-retrieval paradigm involves combining the words in descriptions to construct coherent meanings, inferring the corresponding concept, and mapping it back to words, providing a useful testbed for assessing the way conceptual representations are formed flexibly in large language models. 

As a starting point, we construct description--word pairs from THINGS \citep{hebart_things_2019} and WordNet \citep{fellbaum1998wordnet}, where the description of an object is intended as definitions and hence highly informative of the referent. We use in-context learning paradigm to induce the task routine in the language models. Behavioral assessments across a variety of models show that large language models are able to robustly generate the corresponding lexical items with high accuracy of exact match, given a small number of description--word pairs in the prompt. Representational analysis suggests that the model-constructed conceptual space encodes information about categorical structure and fine-grained object features. Interestingly, models' performances on this reverse dictionary task does not correlates with models' syntactic generalization ability, which may suggest dissociate representation of syntactic knowledge and conceptual knowledge in large language models. Further analysis shows that not only is the models' conceptual inference performance as measured by the reverse-dictionary probe predictive of their general conceptual reasoning ability as evaluated in downstream tasks like commonsense reasoning, incorporating this description$\Rightarrow$word task as prompted examples for language models can induce significant improvements on other reasoning tasks, yielding more human-like behavior.\footnote{Code is available at \url{https://github.com/ningyuxu/tip_of_tongue}.}

\section{Reverse Dictionary for Probing Conceptual Representation}
\label{sec:conceptual-probe}

A common use of language is to talk about things in the mind. 
To achieve this referential goal, listeners have to draw flexible inferences about the concept that a speaker intends to get across from oftentimes a linguistic description of the referent. For example, upon receiving ``a small very thin pancake,'' the listener combines words in this description to derive the underlying meaning, infers the likely referred object concept, and probably retrieves the term ``crepe'' for the referent. This kind of conceptual inference is ubiquitous and necessary to support flexible language understanding and reasoning. To probe the behavioral signatures of flexible inference and representation of concepts in large language models, we re-purpose the classic reverse-dictionary task, i.e. generating the term given a gloss, as a minimal testbed for evaluating language models' capacity for conceptual inference and the structure in the resulted representational space for the inferred concept.

\begin{figure}[t]
\centering
  \includegraphics[width=\linewidth]{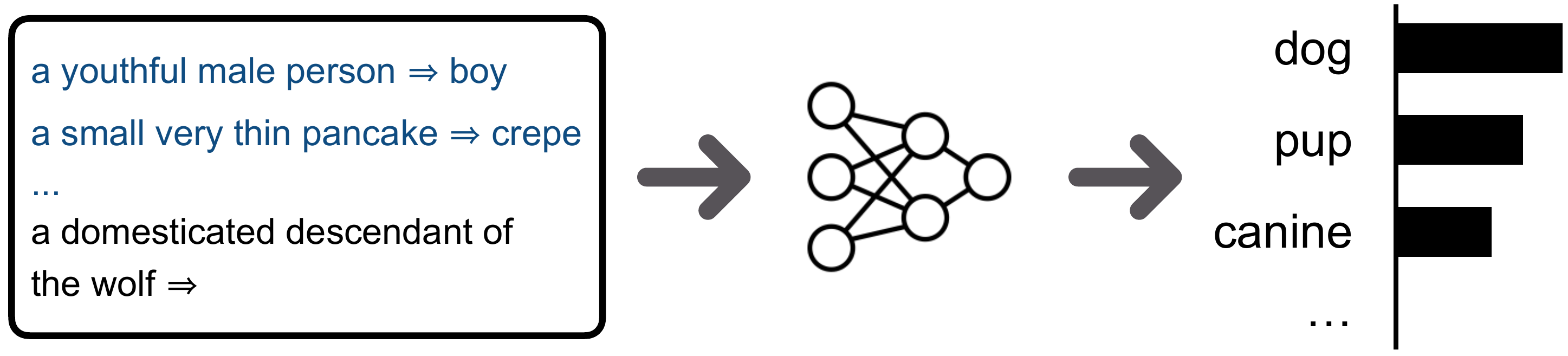}
 \caption{Illustration of the reverse-dictionary probe. A list of $N$ description--word demonstrations is used to prompt an LLM to favorably evoke its conceptual inference capacity. The model generates a word/phrase for the object concept that is described in the query.} \label{fig:rv-dic-demo}
\end{figure}

We take advantage of LLMs' in-context learning ability and derive conceptual representation from them by presenting language models with a small number of demonstrations in a reverse-dictionary format followed by a query description (Figure~\ref{fig:rv-dic-demo}). We compare model-generated completions given the prompt to the name of the object that the query description was originally written for. Specifically, an LLM $\mathcal{M}$ is provided with an input sequence $w_{1:n}$ comprising $n$ tokens, which contains $N$ pairs of descriptions and words as demonstrations $\ell$, along with a query sentence $s$. During inference, the LLM runs them through an embedding layer and $k$ attention layers, encodes the entire sequence into a representation $\mathbf{h}_{n}^{k}$, and then generates the following text based on their probability estimation $p_{\mathcal{M}}\left(\cdot | \ell; s \right)$. We take $\mathbf{h}_{n}^{k}$ as the ``summary'' of the information in the input sequence, which immediately precedes the following predictions that should be in semantic correspondence to the provided description.

We would like to note that while the particular descriptions for prompting and testing LLMs in the following experiments are close to definitions (details of the experimental materials in Section~\ref{sec:conceptual-probe-setup}), they are merely chosen by convenience. Language-based reasoning has to deal with uncertainty, incomplete information, and potentially huge variability in the expressions that people could design to communicate even the same referent. However, the reverse dictionary setup serves as a useful special case to start with. The chosen pairs of concrete nouns and highly informative descriptions create a favorable situation for language models to reveal their competence in meaning representation and concept inference. Models' performances on this special case may indirectly inform their capacity for the challenging case of probabilistic inference.

\subsection{Behavioral Analysis}

\label{sec:conceptual-probe-setup}

We evaluate whether LLMs are able to generate the expected term given an definitional description. We then analyze whether model's performances are robust to variations in the descriptions.

\paragraph{Setup}

We conduct the experiments on 15 open-source Transformer-based \citep{vaswani_attention_2017} LLMs pretrained autoregressively for next-word prediction, including (1) the Falcon models \citep{almazrouei2023falcon, penedo2023refinedweb}, (2) LLaMA \citep{touvron2023llama, touvron2023llama2} models, (3) Mistral 7B \citep{jiang2023mistral}, (4) MPT model \citep{MosaicML2023Introducing}, (5) Phi models \citep{li2023textbooks}, and (6) the Pythia suite \citep{pmlr-v202-biderman23a}.\footnote{We use the LLMs accessible through HuggingFace \citep{wolf2019huggingface}. Additional details can be found in Appendix~\ref{sec:appendix-llms}.} These LLMs vary in architecture, size, and pretraining data, enabling explorative analyses of how these factors might impact model's conceptual inference capacity as measured by the aforementioned reverse-dictionary probe. 

Regarding the experimental materials, we use the description--word pairs primarily sourced from the THINGS database \citep{hebart_things_2019}, which encompasses a broad list of 1,854 concrete and nameable object concepts. We randomly select $N$ word-description pairs as demonstrations and vary $N$ from 1 to 48 to test the impact of the number of demonstrations on LLMs' behavior. To test the robustness of LLMs, we further include in our analysis the corresponding descriptions of these objects in WordNet \citep{fellbaum1998wordnet} and an additional 200 pairs of words and human-written descriptions created by \citet{hill-etal-2016-learning-understand} (referred as Hill200).

We evaluate model performances based on strict exact match across 5 runs. For each concept, we prompt an LLM to generate an answer given a specific description and the arrow symbol ``$\Rightarrow$'', truncate it by ``\texttt{\textbackslash n}'', and then assess whether the resulting output matches the expected word or its synonyms listed in THINGS. We opt for greedy search as our decoding method for a simple and equitable comparison across models.

To interpret language models' performances on the reverse-dictionary task, we construct several control conditions as the baselines: (1) {\textsc{NL}}, where no demonstration is provided and the query is formatted in natural language as ``\texttt{<description> can be called as}''; (2) {\textsc{Mis}}, where each description in the context is paired with a randomly selected word distinct from those in the demonstrations; and (3) {\textsc{Rand}}, where the pairings between descriptions and words undergo complete permutation across the dataset, and the LLMs are evaluated based on matching the randomly-paired word given the query description. We also compare the LLMs' performance with that of the task-specific models reported in previous works \citep{zhang2020multi, yan-etal-2020-bert} for the reverse-dictionary task on the Hill200 dataset.

\begin{figure}[t]
\centering
  \includegraphics[width=7.5cm]{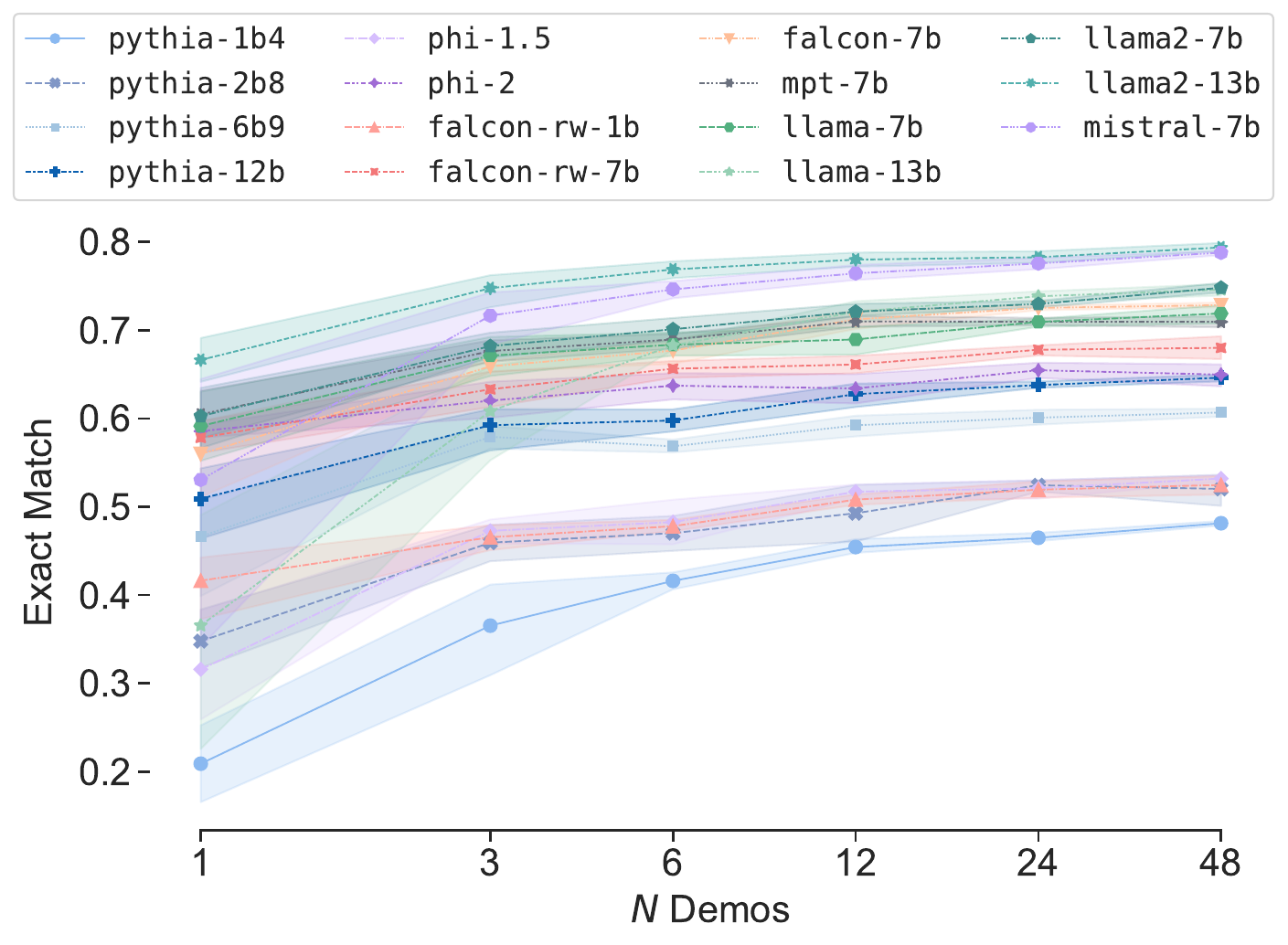}
  \caption{Performance of LLMs in the prompted reverse dictionary task when provided with $N$ description--word pairs. Model performance is measured by exact match between the word/phrase decoded from the model and the name of the specific object for that description. Colored bands denote 95\% confidence intervals.} \label{fig:cp-results}
\end{figure}

\paragraph{Results}

In general, the LLMs we tested here demonstrated great performance in generating the term for the underlying object concept given a definitional description. As shown in Figure~\ref{fig:cp-results}, the average model performance on the description--word pairs from THINGS database notably improves with just three demonstrations and plateaus at approximately 12 to 24 examples. This indicates that a modest number of description--word examples is sufficient to evoke the inference ability. 
Performance comparison with the baselines, especially \textsc{NL}, which on average drops by 25.2\% compared to cases with 24 demonstrations, suggests the benefit of in-context learning in helping reveal the models' capacity for flexible conceptual inference. (see Appendix~\ref{sec:appendix-find-concept-baseline} Table~\ref{tab:concept-inference-baseline}).

There is also notable variability across models. LLMs' performance increases with greater number of parameters ($\rho = 0.76$, see Appendix~\ref{sec:appendix-find-concept-corr-size} Figure~\ref{fig:corr-size}). The performance of \texttt{phi-2} (2.7b), along with the comparison between \texttt{falcon-rw-7b} and \texttt{falcon-7b}, underscores the importance of both scale and quality of pretraining data\footnote{\texttt{falcon-rw-7b} is trained on far less data than \texttt{falcon-7b}.}. 

Beyond the THINGS database, we find that LLMs adeptly adjust to diverse descriptions with minimal performance drop, significantly surpassing previous work \citep{yan-etal-2020-bert} on Hill200 (74\% for LLaMA2-13B compared to 43\% achieved by RoBERTa after explicit training for the reverse-dictionary task, see Appendix~\ref{sec:appendix-desc-impact} Figure~\ref{fig:impact-desc}). We also notice a modest effect of linguistic structure degradation on models' performances when varying degrees of word order permutations are applied to the description, which suggests that the models might be at least sensitive to linguistic structures when combining words into a meaning representation (Model performance decreases by 18\% under full permutation, see Appendix~\ref{sec:appendix-desc-impact} Figure~\ref{fig:impact-shuffle}).

To understand the potential impact of query properties including word frequency, number of word senses, and description length on the model performance, we conducted a correlation analysis based on all 117,659 words in WordNet. We found a moderate overall influence ($\rho = 0.14, 0.08, \textrm{and } 0.12$ respectively, see Appendix~\ref{sec:appendix-query-impact} Figure~\ref{fig:robust-query}). 
Further exploration into the influence of demonstrations is left for future work. 

Taken together, these results indicate the effectiveness and robustness of prompting LLMs to carry out a reverse-dictionary task, laying out the foundation for using this task as a probe for extracting conceptual representation from the model as well as understanding the implications of inference capacity as measured in this task on model's general reasoning ability. Large language models' good performance, as indicated by the high accuracy of exact match, also provides evidence for their general capacity of conceptual inference. 

\subsection{Representation Analysis}

\label{sec:representation}

Human's conceptual representation of objects supports rich inferences about features and properties. 
When thinking of a hedgehog, we also infer that it can be skilled at climbing and digging, typically curls into a tight spiny ball when threatened, and belongs to the category of mammals. These pieces of information can powerfully guide subsequent reasoning. Given large language models' relatively good performances on the reverse-dictionary task in the behavioral analysis, a question naturally arises: does the representational space constructed from the LLMs  encode information about the category structure and fine-grained properties related to the inferred object concept?

\paragraph{Setup}

We run the same set of models as the behavioral analysis on the reverse-dictionary task with 24 demonstrations of description $\Rightarrow$ word. We extract the vector $\mathbf{h}_{n}^{k}$ at the ``$\Rightarrow$''symbol of the query description as the ``summary'' representations of the inferred concept. To probe the structure of the representational space, we conduct two experiments: categorization and feature decoding.

Following \citet{hebart_revealing_2020}, we use the high-level natural categories from the THINGS database as the gold-standard category structure and employ a cross-validated nearest-centroid classifier to assess if the representations derived from conceptual inference are organized in a way that support similarity-based categorization.

\begin{figure*}[t]
\centering
  \includegraphics[width=\linewidth]{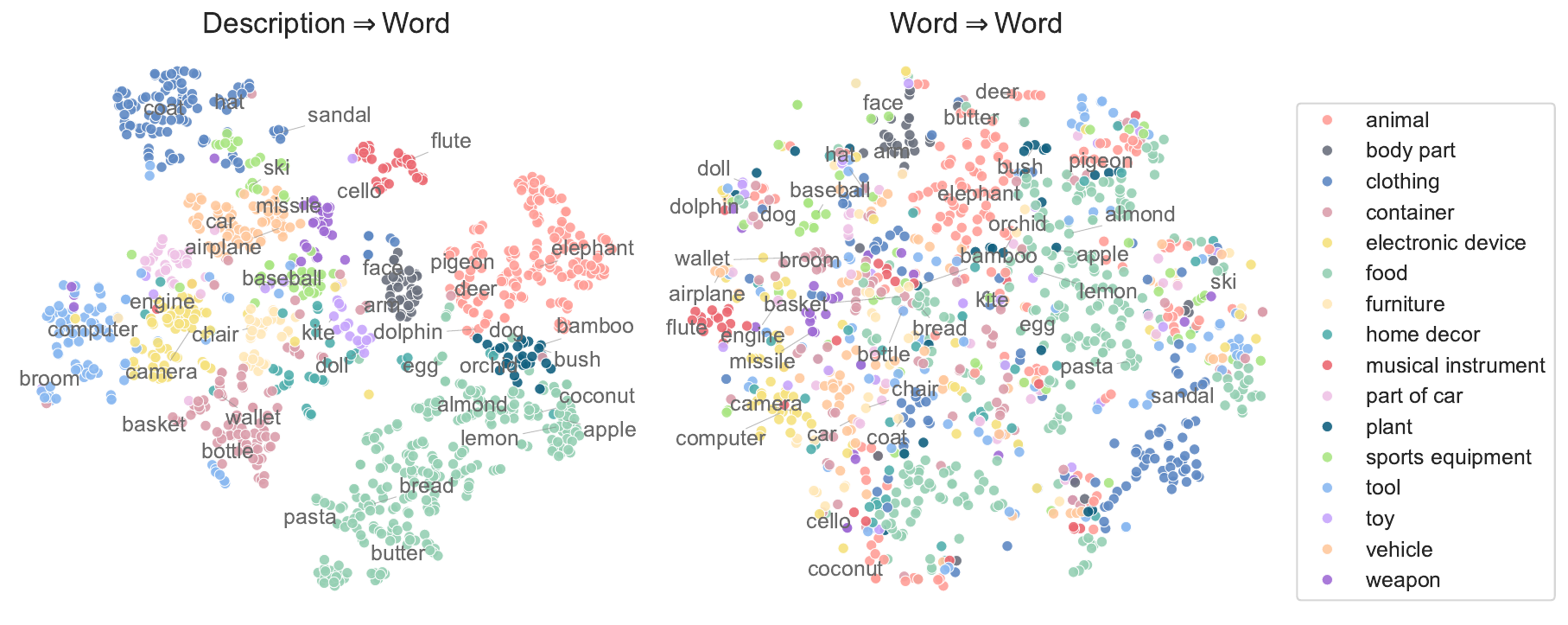}
  \caption{A t-SNE visualization of representations derived from LLaMA2-13B under different task conditions. Representations are extracted at the ``$\Rightarrow$'' symbol. Category assignments are based on the THINGS data.
  } \label{fig:category-space-visualization-comparison}
\end{figure*}

We then explore whether model representations encode information about fine-grained features associated with the concepts. We use the XCSLB dataset \citep{misra2022property}, which comprises 3,645 human-generated binary descriptive features, such as \textit{live under water} (true for \textsc{jellyfish} and false for \textsc{butterfly}). We train feature-specific logistic regression models to predict the feature value for the test items and report the average $F_{1}$ scores and area under the curve (AUC) in 10-fold cross-validation, similar to the evaluation procedure in \citet{zheng2018revealing}.

In comparison, we run the same categorization and feature decoding experiments with baseline representations, including static word embeddings and LLM representations that are contextually formed but not in the context of concept inference: (1) \textsc{fastText}, the static word embeddings trained using fastText on Common Crawl and Wikipedia \citep{grave-etal-2018-learning}, which is commonly used to investigate the knowledge derived from language data, (2) \textsc{SPoSE} \citep{hebart_revealing_2020}, an embedding that supports stable prediction of human similarity judgments over the concepts in THINGS as well as the categorization behavior, (3) {\textsc{Word}}, the word representations derived through inputting the word to LLMs, (4) \textsc{Description}, the representation of the description LLMs form before seeing the delimiter and (5) \textsc{W2W}, where we give $N$ demonstrations in the format of ``\texttt{<Word> $\Rightarrow$ <Word>}'' to LLMs to elicit prediction of the same word as in the reverse-dictionary case, but successful prompt completion does not necessarily engage in reasoning about the concept underlying the input word. We also include representations derived from the baselines outlined in the previous subsection ({\textsc{Mis}} and {\textsc{NL}}).

\paragraph{Results}

The summary representation extracted from LLMs generally supports similarity-based categorization, achieving an average performance at around 90\% and surpassing all the baselines including \textsc{fastText} (78\%) and \textsc{SPoSE} (86\%). Crucially, the contextualized representation formed in the word$\Rightarrow$word input repetition task ({\textsc{W2W}}) yields worse performance (ranging from about 60\% to 85\%) compared to the description$\Rightarrow$word task, and the difference in the strutural alignment with human-annotated category space is qualitatively notable when visualizing the representational space in lower dimensions in Figure~\ref{fig:category-space-visualization-comparison}. This suggests that while LLMs have learned richly-structured word representations---at least for concrete nouns---that support categorization to some degree, the representations that the models formed given the reverse-dictionary probe produce a more structurally-aligned representational space for the underlying concepts. This is also evidenced by the subpar performance of other baselines including \textsc{Word}, \textsc{Description}, \textsc{NL} and \textsc{Mis} (see Appendix~\ref{sec:appendix-probing-structure} Table~\ref{tab:categorization}), which shows that simply providing the descriptions or words alone to LLMs does not necessarily gives rise to a representational space that structurally aligns with human-like object categories as closely as the ones extracted from the reverse-dictionary probe.

In addition to the great performance in object categorization, we find that the representations that LLMs construct contain decodable information about fine-grained features. On average, model representations achieve a $F_{1}$ score of approximately 80\% and an AUC of around 96\% in terms of mapping representations to binary features annotated in XCSLB. Across models, feature decoding performances are higher for taxonomic and encyclopedic features over visual and perceptual ones (Detailed results are shown in Appendix~\ref{sec:appendix-probing-feature} Table~\ref{tab:feature-prediction} and Figure~\ref{fig:xcslb-all}). This might stem from the exclusive reliance on language data in the model training procedure. 
We also note that certain baselines, especially \textsc{W2W}, also perform relatively well in decoding fine-grained object properties despite less compelling performance in the categorization experiment. We conjecture that while the word representations of LLMs might not be structured in such a way that readily supports simple similarity-based categorization, they may still encode fine-grained distinctions among different lexical concepts that enables effective learning of binary feature classifiers.

\section{Implications of Conceptual Inference on Models' Generalization Behaviors}

The reverse-dictionary probe as introduced in Section~\ref{sec:conceptual-probe} measures LLMs' competence for conceptual inference via a specific test case. One might wonder whether results from this minimal test case reveal any meaningful behavioral signatures about models' general language-based reasoning ability.

There are reasons to think of this reverse-dictionary task as not just yet another new thing that LLMs can do, but a useful and targeted probe into the model's capacity to perform a canonical computation that underlies various complex reasoning behaviors. To explore this idea, we conduct three experiments to study the relationship between model's conceptual inference capacity, as measured by the reverse-dictionary probe, and model's generalization behaviors.

\subsection{Conceptual Inference Ability Predicts Commonsense Reasoning Performance}\label{sec:concept-inference-commonsense-reasoning}

\label{sec:role-indicator}

\begin{figure}[t]
\centering
  \includegraphics[width=\linewidth]{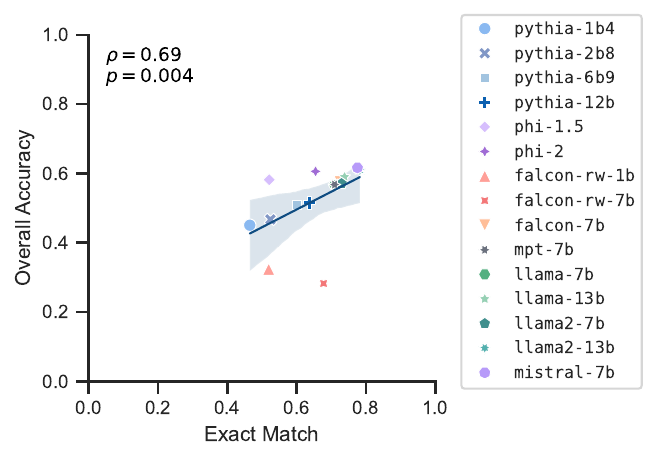}
  \caption{Correlation between LLMs' overall performance averaged across different reasoning tasks and their average conceptual inference performance in the reverse dictionary task with 24 demonstrations provided. 
  } \label{fig:corr-reasoning}
\end{figure}

\paragraph{Setup}

We conduct a correlation analysis to examine the relationship between conceptual inference and the general commonsense reasoning abilities of LLMs. We take widely-used benchmarks to evaluate LLMs' general knowledge and reasoning ability, including CommonsenseQA \citep[CSQA;][]{talmor-etal-2019-commonsenseqa}, ARC easy (ARC-E) and challenge \citep[ARC-C;][]{clark2018think}, OpenBookQA \citep{mihaylov-etal-2018-suit}, PIQA \citep{bisk_piqa_2020}, SIQA \citep{sap-etal-2019-social}, Hellaswag \citep{zellers-etal-2019-hellaswag} and BoolQ \citep{clark-etal-2019-boolq}. The tasks in these benchmarks are all formatted as multiple-choice questions, where a model is typically presented with a query (e.g., ``\textit{Where is a bald eagle safe?}'') and evaluated by their accuracy in ranking the correct answer (e.g., ``\textit{wildlife refuge}'') with the highest probability among alternatives (e.g., ``\textit{in washington}'' and ``\textit{open country}''). 

We use the test sets of each task for evaluation if publicly available; otherwise we resort to the development set. LLMs are evaluated in a zero-shot manner through natural language prompt templates, with the score of each answer computed as the sum of log-likelihoods LLMs assign to it (see Appendix~\ref{sec:appendix-corr-general-evaluation} for details).

\paragraph{Results}

Figure~\ref{fig:corr-reasoning} shows a significant correlation between LLMs' conceptual inference ability, as probed through the reverse-dictionary task, and their average performance across various commonsense reasoning tasks (see Appendix~\ref{sec:appendix-corr-general-results} Figure~\ref{fig:corr-reasoning-all-tasks} for correlation results on each task). 
These findings suggest that the degree to which a model can flexibly engage with concept inference, even as measured in such a constrained domain (concepts of concrete objects), might account for the observed cross-model differences in general reasoning capacity.

\subsection{Relationship between Conceptual Inference and Syntactic Generalization}\label{sec:conceptual-inference-vs-syntactic-generalization}

Meaning composition entails combining words in a way that conforms to the syntactic structure \citep{partee1984compositionality}, but do LLMs rely on syntactic knowledge for constructing conceptual representations? Experiment 2 investigates the relationship between conceptual inference and syntactic generalization in LLMs by comparing their performance probed by the reverse-dictionary task with that in targeted syntactic evaluations. 

\paragraph{Setup}

We use two benchmarks for evaluating models' syntactic generalization: SyntaxGym \citep{hu-etal-2020-systematic,gauthier-etal-2020-syntaxgym} and the Benchmark of Linguistic Minimal Pairs \citep[BLiMP;][]{warstadt-etal-2020-blimp-benchmark}, which cover a wide range of linguistic phenomena. Both benchmarks construct controlled English stimuli to assess a model's syntactic generalization behavior. The evaluation paradigm of SyntaxGym is based on whether a language model generates human-like differentiable expectations about upcoming linguistic materials given the structural information in the prefix. BLiMP's paradigm compares a model's likelihood assignments between a well-formed sentence and minimally different ungrammatical counterpart.
We prepend a \texttt{[BOS]} token to each sentence before inputting it to the model. 
We report the accuracy averaged across the test suites for both benchmarks. 
Accuracy scores for particular test suites can be found in Figure~\ref{fig:syntax-details} in Appendix~\ref{sec:appendix-syntax}.

\paragraph{Results}

\begin{figure}[t]
\centering
  \includegraphics[width=\linewidth]{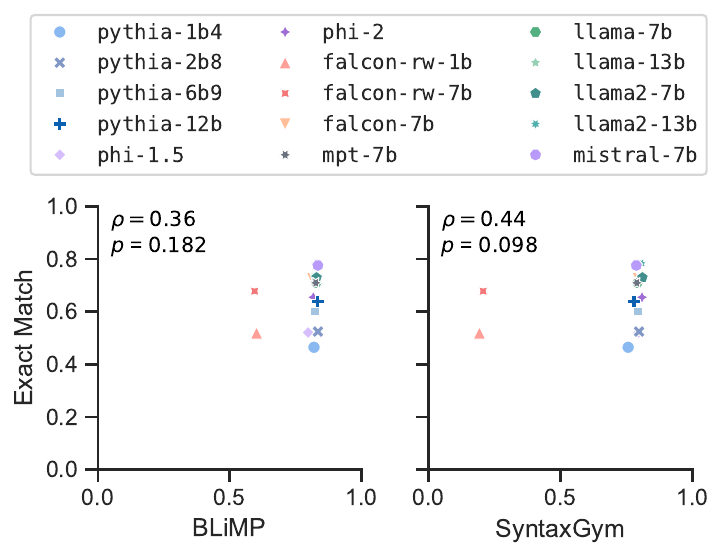}
  \caption{Correlation between the LLMs' syntactic generalization ability, as measured by BLiMP (Left) and SyntaxGym (Right), and their average performance in the conceptual inference task with 24 demonstrations.} \label{fig:corr-syntax}
\end{figure}

While large language models exhibit significant variability in their conceptual inference ability as measured by the reverse-dictionary task in Section~\ref{sec:conceptual-probe}, the vast majority of the models tested here perform similarly well on the syntactic generalization benchmarks  (Figure~\ref{fig:corr-syntax}). The \texttt{falcon-rw} models, trained exclusively on web data \citep{penedo2023refinedweb}, are the outliers that achieve comparatively lower performance in syntactic evaluation, potentially because the web data contains a lot of noises and language production errors. This result also suggests that the observed correlation between a model's performance on the reverse-dictionary task and its performance on other reasoning tasks are not an epiphenomenon of a powerful model being good at every tasks. From a different perspective, a model's syntactic generalization ability does not seem to improve along with an increased capacity for conceptual inference. This raises a puzzle for future work about the relationship between linguistic generalization and conceptual reasoning in large language models.

\subsection{Generalizing Reverse Dictionary to Commonsense Reasoning}\label{sec:generalize-to-reasoning}

Our final experiment investigates whether guiding LLMs for conceptual inference may facilitate the models in approaching tasks that involves reasoning about items congruent with the meaning of a given phrase, even if the query task may be substantially different from the prompt examples in terms of the content of the involved reasoning process. We focus on commonsense reasoning and use ProtoQA \citep{boratko-etal-2020-protoqa} for experiment. ProtoQA presents prototypical situations with many plausible answers, with some more typical than others, e.g., ``\textit{Name something that you might forget in a hotel room}.'' We analyze the impact of conceptual inference on LLMs' behavior by comparing their performance with that in zero-shot scenarios and under different prompts.


\paragraph{Setup}

We use the development set of ProtoQA for evaluation as the answers to the test sets are not publicly available. We follow the evaluation protocol in the original paper, where diverse answers sampled from LLMs are compared with human-generated ones through the criteria of exact match and matching with synonyms in WordNet. We report Max Answers@$k$ and Max Incorrect@$k$, where Max Answers@$k$ restricts the total number of answers to $k$, and Max Incorrect@$k$ halts after $k$ unmatched answers are provided (Additional details can be found in Appendix~\ref{sec:appendix-generalize-evaluation}). To evaluate the influence of conceptual inference on LLMs' behavior, as in Section~\ref{sec:conceptual-probe}, we provide the LLM with an input sequence $w_{1:n}$ that comprises $N$ description$\Rightarrow$word pairs $\ell$ and a query sentence $s$ drawn from the evaluation dataset. We then compare the performance when $N \in \left\{1, 12, 24\right\}$ demonstrations are given and incorporate the \textsc{NL} baseline, where we use the natural language prompt templates modified for next-word prediction.

\begin{figure}[t]
\centering
  \includegraphics[width=\linewidth]{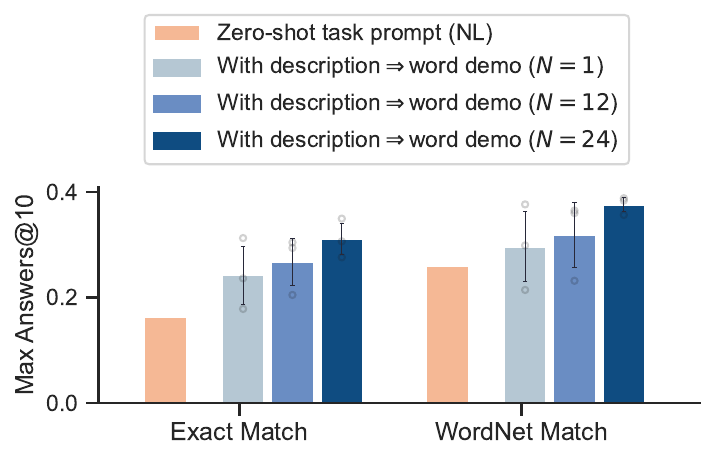}
  \caption{Performance of LLaMA2-13B in ProtoQA evaluated by Max Answers@10 under the natural language task prompt (\textsc{NL}) and formatted reverse dictionary prompt with $N$ description$\Rightarrow$word demonstrations. 
  } \label{fig:protoqa-causal}
\end{figure}

\paragraph{Results}

The performance of LLMs in ProtoQA improves given the reverse-dictionary demonstrations (Figure~\ref{fig:protoqa-causal}), generally surpassing the zero-shot setting where task-specific natural language templates are used (Detailed results are shown in Appendix~\ref{sec:appendix-generalize-results} Table~\ref{tab:protoqa-causal}). While LLMs exhibit the ability to generate reasonable answers when prompted with \textsc{NL}, the responses are typically verbose and occasionally contain irrelevant information. When guided by reverse-dictionary examples, LLMs tend to produce precise answers that align more closely with human-generated answers, without any modification of the original questions (see Table~\ref{tab:protoqa-llm-answers} in Appendix~\ref{sec:appendix-generalize-results} for examples of LLM-generated answers). While we do not claim that the reverse dictionary demonstrations work better than other task-specific prompts or hand-designed templates that align with the next-word prediction pretraining objective, the observed generalization ability of LLMs suggests that 
the reverse-dictionary demonstrations can guide the LLMs to go beyond a specific task construal and learn to construct useful representations for commonsense reasoning.

\section{Related Work}

The impressive performance of LLMs across various language comprehension benchmarks has sparked debates about conceptual representations in these models \citep{bender-koller-2020-climbing, piantadosi2022meaning, mitchell_debate_2023} as well as their relevance to understanding the human mind \citep{binz2023turning, frank_openly_2023, hardy2023large}. Previous work suggests that LLMs demonstrate human-like behavior in some aspects of reasoning \citep{webb_emergent_2023, hagendorff_human-like_2023, dasgupta2022language, han2024inductive} and semantic structure \citep{hansen2022semantic, marjieh2022predicting}, but these models tend to be overly sensitive to contextual variations \citep{binz_using_2023, wu2023reasoning, suresh-etal-2023-conceptual}. Analyses of their representations demonstrate their effectiveness in encoding world knowledge \citep{da-kasai-2019-cracking, forbes2019neural} and dynamically forming world state representations \citep{li2023emergent,yamakoshi-etal-2023-causal, li-etal-2021-implicit}. Research has also looked into model's ability to reason about and make inductive inferences about object properties \citep{misra-etal-2023-comps,han2024inductive}.

Our work complements existing approaches by focusing on a canonical example of conceptual inference: naming an intended referent that is described indirectly. A special case of this general inference problem, reverse dictionary, has been a familiar problem in the NLP community, and approached with trained or fine-tuned task-specific neural network models \citep{hill-etal-2016-learning-understand, zhang2020multi, yan-etal-2020-bert,siddique_reverse_2023}. We combine this classic task with a novel dataset of object concepts (THINGS) to develop a minimal testbed for probing conceptual representations in large language models, adding new kinds of evidence to the threads of research on evaluating language models' reasoning capacity.

\section{Conclusion}

Concepts bridge the thoughts and the words. 
Here we take the classic reverse dictionary task to probe the conceptual inference capacity in large language models.
Given a few description--word pairs, LLMs effectively learn to infer concepts from complex linguistic descriptions. 
The contextually-formed representational space in the models structurally aligns with the space of object categories and maintains fine-grained distinctions across individual concepts along various feature dimensions. 
To the degree that large language models demonstrate promising behaviors in a minimal case of conceptual inference, our approach may open new questions about the nature and limit of their learned capacity for meaning representation.

\section*{Limitations}

Compositionality in natural language is complex and intricate. 
While the reverse dictionary task in principle involves combining word representation into a conceptual representation, the design of this study does not afford an in-depth analysis of phrase-level meaning composition. In addition, this work does not provide a mechanistic explanation of how LLMs achieve the ability to do reverse dictionary task after being prompted with a few demonstrations. 

Our experimental materials use definitional descriptions about concrete objects. Although this is an intentional choice, we note here that it might constrain how well the experimental results can generalize to a general case of probabilistic inference. While our main research objective is not about building a reverse dictionary, wider range of words and terms, including different part-of-speech categories and domains, are needed to critically assess the potential of turning a prompted LLM into a ready-to-go reverse dictionary application. On the side of understanding conceptual representations in LLMs, diverse domains of concepts are also relevant for painting a fuller picture of the models' competence and potential limitations.


\bibliography{anthology, custom}

\begin{thebibliography}{70}
\expandafter\ifx\csname natexlab\endcsname\relax\def\natexlab#1{#1}\fi

\bibitem[{Almazrouei et~al.(2023)Almazrouei, Alobeidli, Alshamsi, Cappelli, Cojocaru, Debbah, Goffinet, Hesslow, Launay, Malartic et~al.}]{almazrouei2023falcon}
Ebtesam Almazrouei, Hamza Alobeidli, Abdulaziz Alshamsi, Alessandro Cappelli, Ruxandra Cojocaru, M{\'e}rouane Debbah, {\'E}tienne Goffinet, Daniel Hesslow, Julien Launay, Quentin Malartic, et~al. 2023.
\newblock \href {https://arxiv.org/abs/2311.16867} {The falcon series of open language models}.
\newblock \emph{arXiv preprint arXiv:2311.16867}.

\bibitem[{Bender and Koller(2020)}]{bender-koller-2020-climbing}
Emily~M. Bender and Alexander Koller. 2020.
\newblock \href {https://doi.org/10.18653/v1/2020.acl-main.463} {Climbing towards {NLU}: {On} meaning, form, and understanding in the age of data}.
\newblock In \emph{Proceedings of the 58th Annual Meeting of the Association for Computational Linguistics}, pages 5185--5198, Online. Association for Computational Linguistics.

\bibitem[{Biderman et~al.(2023)Biderman, Schoelkopf, Anthony, Bradley, O'Brien, Hallahan, Khan, Purohit, Prashanth, Raff, Skowron, Sutawika, and Van Der~Wal}]{pmlr-v202-biderman23a}
Stella Biderman, Hailey Schoelkopf, Quentin~Gregory Anthony, Herbie Bradley, Kyle O'Brien, Eric Hallahan, Mohammad~Aflah Khan, Shivanshu Purohit, Usvsn~Sai Prashanth, Edward Raff, Aviya Skowron, Lintang Sutawika, and Oskar Van Der~Wal. 2023.
\newblock \href {https://proceedings.mlr.press/v202/biderman23a.html} {Pythia: A suite for analyzing large language models across training and scaling}.
\newblock In \emph{Proceedings of the 40th International Conference on Machine Learning}, volume 202 of \emph{Proceedings of Machine Learning Research}, pages 2397--2430. PMLR.

\bibitem[{Binz and Schulz(2023{\natexlab{a}})}]{binz2023turning}
Marcel Binz and Eric Schulz. 2023{\natexlab{a}}.
\newblock \href {https://arxiv.org/abs/2306.03917} {Turning large language models into cognitive models}.
\newblock \emph{arXiv preprint arXiv:2306.03917}.

\bibitem[{Binz and Schulz(2023{\natexlab{b}})}]{binz_using_2023}
Marcel Binz and Eric Schulz. 2023{\natexlab{b}}.
\newblock \href {https://doi.org/10.1073/pnas.2218523120} {Using cognitive psychology to understand gpt-3}.
\newblock \emph{Proceedings of the National Academy of Sciences}, 120(6):e2218523120.

\bibitem[{Bisk et~al.(2020)Bisk, Zellers, Le~bras, Gao, and Choi}]{bisk_piqa_2020}
Yonatan Bisk, Rowan Zellers, Ronan Le~bras, Jianfeng Gao, and Yejin Choi. 2020.
\newblock \href {https://doi.org/10.1609/aaai.v34i05.6239} {Piqa: Reasoning about physical commonsense in natural language}.
\newblock \emph{Proceedings of the AAAI Conference on Artificial Intelligence}, 34(05):7432--7439.

\bibitem[{Boratko et~al.(2020)Boratko, Li, O{'}Gorman, Das, Le, and McCallum}]{boratko-etal-2020-protoqa}
Michael Boratko, Xiang Li, Tim O{'}Gorman, Rajarshi Das, Dan Le, and Andrew McCallum. 2020.
\newblock \href {https://doi.org/10.18653/v1/2020.emnlp-main.85} {{P}roto{QA}: A question answering dataset for prototypical common-sense reasoning}.
\newblock In \emph{Proceedings of the 2020 Conference on Empirical Methods in Natural Language Processing (EMNLP)}, pages 1122--1136, Online. Association for Computational Linguistics.

\bibitem[{Bubeck et~al.(2023)Bubeck, Chandrasekaran, Eldan, Gehrke, Horvitz, Kamar, Lee, Lee, Li, Lundberg et~al.}]{bubeck2023sparks}
S{\'e}bastien Bubeck, Varun Chandrasekaran, Ronen Eldan, Johannes Gehrke, Eric Horvitz, Ece Kamar, Peter Lee, Yin~Tat Lee, Yuanzhi Li, Scott Lundberg, et~al. 2023.
\newblock \href {https://arxiv.org/abs/2303.12712} {Sparks of artificial general intelligence: Early experiments with gpt-4}.
\newblock \emph{arXiv preprint arXiv:2303.12712}.

\bibitem[{Casas et~al.(2019)Casas, Hernández-Fernández, Català, i~Cancho, and Baixeries}]{CASAS201919}
Bernardino Casas, Antoni Hernández-Fernández, Neus Català, Ramon~Ferrer i~Cancho, and Jaume Baixeries. 2019.
\newblock \href {https://doi.org/https://doi.org/10.1016/j.csl.2019.03.007} {Polysemy and brevity versus frequency in language}.
\newblock \emph{Computer Speech \& Language}, 58:19--50.

\bibitem[{Chang and Bergen(2022)}]{chang-bergen-2022-word}
Tyler~A. Chang and Benjamin~K. Bergen. 2022.
\newblock \href {https://doi.org/10.1162/tacl_a_00444} {Word acquisition in neural language models}.
\newblock \emph{Transactions of the Association for Computational Linguistics}, 10:1--16.

\bibitem[{Clark et~al.(2019)Clark, Lee, Chang, Kwiatkowski, Collins, and Toutanova}]{clark-etal-2019-boolq}
Christopher Clark, Kenton Lee, Ming-Wei Chang, Tom Kwiatkowski, Michael Collins, and Kristina Toutanova. 2019.
\newblock \href {https://doi.org/10.18653/v1/N19-1300} {{B}ool{Q}: Exploring the surprising difficulty of natural yes/no questions}.
\newblock In \emph{Proceedings of the 2019 Conference of the North {A}merican Chapter of the Association for Computational Linguistics: Human Language Technologies, Volume 1 (Long and Short Papers)}, pages 2924--2936, Minneapolis, Minnesota. Association for Computational Linguistics.

\bibitem[{Clark et~al.(2018)Clark, Cowhey, Etzioni, Khot, Sabharwal, Schoenick, and Tafjord}]{clark2018think}
Peter Clark, Isaac Cowhey, Oren Etzioni, Tushar Khot, Ashish Sabharwal, Carissa Schoenick, and Oyvind Tafjord. 2018.
\newblock \href {https://arxiv.org/abs/1803.05457} {Think you have solved question answering? try arc, the ai2 reasoning challenge}.
\newblock \emph{arXiv preprint arXiv:1803.05457}.

\bibitem[{Da and Kasai(2019)}]{da-kasai-2019-cracking}
Jeff Da and Jungo Kasai. 2019.
\newblock \href {https://doi.org/10.18653/v1/D19-6001} {Cracking the contextual commonsense code: Understanding commonsense reasoning aptitude of deep contextual representations}.
\newblock In \emph{Proceedings of the First Workshop on Commonsense Inference in Natural Language Processing}, pages 1--12, Hong Kong, China. Association for Computational Linguistics.

\bibitem[{Dasgupta et~al.(2022)Dasgupta, Lampinen, Chan, Creswell, Kumaran, McClelland, and Hill}]{dasgupta2022language}
Ishita Dasgupta, Andrew~K Lampinen, Stephanie~CY Chan, Antonia Creswell, Dharshan Kumaran, James~L McClelland, and Felix Hill. 2022.
\newblock \href {https://arxiv.org/abs/2207.07051} {Language models show human-like content effects on reasoning}.
\newblock \emph{arXiv preprint arXiv:2207.07051}.

\bibitem[{Devereux et~al.(2014)Devereux, Tyler, Geertzen, and Randall}]{devereux_centre_2014}
Barry~J. Devereux, Lorraine~K. Tyler, Jeroen Geertzen, and Billi Randall. 2014.
\newblock \href {https://doi.org/10.3758/s13428-013-0420-4} {The {Centre} for {Speech}, {Language} and the {Brain} ({CSLB}) concept property norms}.
\newblock \emph{Behavior Research Methods}, 46(4):1119--1127.

\bibitem[{Fellbaum(1998)}]{fellbaum1998wordnet}
Christiane Fellbaum. 1998.
\newblock \emph{WordNet: An electronic lexical database}.
\newblock MIT press.

\bibitem[{Forbes et~al.(2019)Forbes, Holtzman, and Choi}]{forbes2019neural}
Maxwell Forbes, Ari Holtzman, and Yejin Choi. 2019.
\newblock \href {https://arxiv.org/abs/1908.02899} {Do neural language representations learn physical commonsense?}
\newblock \emph{arXiv preprint arXiv:1908.02899}.

\bibitem[{Frank(2023)}]{frank_openly_2023}
Michael~C. Frank. 2023.
\newblock \href {https://doi.org/10.1038/s41562-023-01732-4} {Openly accessible {LLMs} can help us to understand human cognition}.
\newblock \emph{Nature Human Behaviour}, 7(11):1825--1827.

\bibitem[{Gauthier et~al.(2020)Gauthier, Hu, Wilcox, Qian, and Levy}]{gauthier-etal-2020-syntaxgym}
Jon Gauthier, Jennifer Hu, Ethan Wilcox, Peng Qian, and Roger Levy. 2020.
\newblock \href {https://doi.org/10.18653/v1/2020.acl-demos.10} {{S}yntax{G}ym: An online platform for targeted evaluation of language models}.
\newblock In \emph{Proceedings of the 58th Annual Meeting of the Association for Computational Linguistics: System Demonstrations}, pages 70--76, Online. Association for Computational Linguistics.

\bibitem[{Geiger et~al.(2023)Geiger, Wu, Potts, Icard, and Goodman}]{geiger2023finding}
Atticus Geiger, Zhengxuan Wu, Christopher Potts, Thomas Icard, and Noah~D Goodman. 2023.
\newblock \href {https://arxiv.org/abs/2303.02536} {Finding alignments between interpretable causal variables and distributed neural representations}.
\newblock \emph{arXiv preprint arXiv:2303.02536}.

\bibitem[{Grave et~al.(2018)Grave, Bojanowski, Gupta, Joulin, and Mikolov}]{grave-etal-2018-learning}
Edouard Grave, Piotr Bojanowski, Prakhar Gupta, Armand Joulin, and Tomas Mikolov. 2018.
\newblock \href {https://aclanthology.org/L18-1550} {Learning word vectors for 157 languages}.
\newblock In \emph{Proceedings of the Eleventh International Conference on Language Resources and Evaluation ({LREC} 2018)}, Miyazaki, Japan. European Language Resources Association (ELRA).

\bibitem[{Gurnee and Tegmark(2024)}]{gurnee2023language}
Wes Gurnee and Max Tegmark. 2024.
\newblock \href {https://openreview.net/forum?id=jE8xbmvFin} {Language models represent space and time}.
\newblock In \emph{The Twelfth International Conference on Learning Representations}.

\bibitem[{Hagendorff et~al.(2023)Hagendorff, Fabi, and Kosinski}]{hagendorff_human-like_2023}
Thilo Hagendorff, Sarah Fabi, and Michal Kosinski. 2023.
\newblock \href {https://doi.org/10.1038/s43588-023-00527-x} {Human-like intuitive behavior and reasoning biases emerged in large language models but disappeared in {ChatGPT}}.
\newblock \emph{Nature Computational Science}, 3(10):833--838.

\bibitem[{Han et~al.(2024)Han, Ransom, Perfors, and Kemp}]{han2024inductive}
Simon~Jerome Han, Keith~J. Ransom, Andrew Perfors, and Charles Kemp. 2024.
\newblock \href {https://doi.org/https://doi.org/10.1016/j.cogsys.2023.101155} {Inductive reasoning in humans and large language models}.
\newblock \emph{Cognitive Systems Research}, 83:101155.

\bibitem[{Hansen and Hebart(2022)}]{hansen2022semantic}
Hannes Hansen and Martin~N Hebart. 2022.
\newblock \href {https://escholarship.org/uc/item/44s454ng} {Semantic features of object concepts generated with {GPT}-3}.
\newblock In \emph{Proceedings of the Annual Meeting of the Cognitive Science Society}, volume~44.

\bibitem[{Hardy et~al.(2023)Hardy, Sucholutsky, Thompson, and Griffiths}]{hardy2023large}
Mathew Hardy, Ilia Sucholutsky, Bill Thompson, and Tom Griffiths. 2023.
\newblock \href {https://escholarship.org/uc/item/6dp9k2gz} {Large language models meet cognitive science: Llms as tools, models, and participants}.
\newblock In \emph{Proceedings of the annual meeting of the cognitive science society}, volume~45.

\bibitem[{Hebart et~al.(2019)Hebart, Dickter, Kidder, Kwok, Corriveau, Van~Wicklin, and Baker}]{hebart_things_2019}
Martin~N. Hebart, Adam~H. Dickter, Alexis Kidder, Wan~Y. Kwok, Anna Corriveau, Caitlin Van~Wicklin, and Chris~I. Baker. 2019.
\newblock \href {https://doi.org/10.1371/journal.pone.0223792} {Things: A database of 1,854 object concepts and more than 26,000 naturalistic object images}.
\newblock \emph{PLOS ONE}, 14(10):1--24.

\bibitem[{Hebart et~al.(2020)Hebart, Zheng, Pereira, and Baker}]{hebart_revealing_2020}
Martin~N. Hebart, Charles~Y. Zheng, Francisco Pereira, and Chris~I. Baker. 2020.
\newblock \href {https://doi.org/10.1038/s41562-020-00951-3} {Revealing the multidimensional mental representations of natural objects underlying human similarity judgements}.
\newblock \emph{Nature Human Behaviour}, 4(11):1173--1185.

\bibitem[{Hill et~al.(2016)Hill, Cho, Korhonen, and Bengio}]{hill-etal-2016-learning-understand}
Felix Hill, Kyunghyun Cho, Anna Korhonen, and Yoshua Bengio. 2016.
\newblock \href {https://doi.org/10.1162/tacl_a_00080} {Learning to understand phrases by embedding the dictionary}.
\newblock \emph{Transactions of the Association for Computational Linguistics}, 4:17--30.

\bibitem[{Holtzman et~al.(2020)Holtzman, Buys, Du, Forbes, and Choi}]{Holtzman2020The}
Ari Holtzman, Jan Buys, Li~Du, Maxwell Forbes, and Yejin Choi. 2020.
\newblock \href {https://openreview.net/forum?id=rygGQyrFvH} {The curious case of neural text degeneration}.
\newblock In \emph{International Conference on Learning Representations}.

\bibitem[{Hu et~al.(2020)Hu, Gauthier, Qian, Wilcox, and Levy}]{hu-etal-2020-systematic}
Jennifer Hu, Jon Gauthier, Peng Qian, Ethan Wilcox, and Roger Levy. 2020.
\newblock \href {https://doi.org/10.18653/v1/2020.acl-main.158} {A systematic assessment of syntactic generalization in neural language models}.
\newblock In \emph{Proceedings of the 58th Annual Meeting of the Association for Computational Linguistics}, pages 1725--1744, Online. Association for Computational Linguistics.

\bibitem[{Jiang et~al.(2023)Jiang, Sablayrolles, Mensch, Bamford, Chaplot, Casas, Bressand, Lengyel, Lample, Saulnier et~al.}]{jiang2023mistral}
Albert~Q Jiang, Alexandre Sablayrolles, Arthur Mensch, Chris Bamford, Devendra~Singh Chaplot, Diego de~las Casas, Florian Bressand, Gianna Lengyel, Guillaume Lample, Lucile Saulnier, et~al. 2023.
\newblock \href {https://arxiv.org/abs/2310.06825} {Mistral 7b}.
\newblock \emph{arXiv preprint arXiv:2310.06825}.

\bibitem[{Kandpal et~al.(2023)Kandpal, Deng, Roberts, Wallace, and Raffel}]{pmlr-v202-kandpal23a}
Nikhil Kandpal, Haikang Deng, Adam Roberts, Eric Wallace, and Colin Raffel. 2023.
\newblock \href {https://proceedings.mlr.press/v202/kandpal23a.html} {Large language models struggle to learn long-tail knowledge}.
\newblock In \emph{Proceedings of the 40th International Conference on Machine Learning}, volume 202 of \emph{Proceedings of Machine Learning Research}, pages 15696--15707. PMLR.

\bibitem[{Leech et~al.(1994)Leech, Garside, and Bryant}]{leech-etal-1994-claws4}
Geoffrey Leech, Roger Garside, and Michael Bryant. 1994.
\newblock \href {https://aclanthology.org/C94-1103} {{CLAWS}4: The tagging of the {B}ritish {N}ational {C}orpus}.
\newblock In \emph{{COLING} 1994 Volume 1: The 15th {I}nternational {C}onference on {C}omputational {L}inguistics}, Kyoto, Japan.

\bibitem[{Li et~al.(2021)Li, Nye, and Andreas}]{li-etal-2021-implicit}
Belinda~Z. Li, Maxwell Nye, and Jacob Andreas. 2021.
\newblock \href {https://doi.org/10.18653/v1/2021.acl-long.143} {Implicit representations of meaning in neural language models}.
\newblock In \emph{Proceedings of the 59th Annual Meeting of the Association for Computational Linguistics and the 11th International Joint Conference on Natural Language Processing (Volume 1: Long Papers)}, pages 1813--1827, Online. Association for Computational Linguistics.

\bibitem[{Li et~al.(2023{\natexlab{a}})Li, Hopkins, Bau, Vi{\'e}gas, Pfister, and Wattenberg}]{li2023emergent}
Kenneth Li, Aspen~K Hopkins, David Bau, Fernanda Vi{\'e}gas, Hanspeter Pfister, and Martin Wattenberg. 2023{\natexlab{a}}.
\newblock \href {https://openreview.net/forum?id=DeG07_TcZvT} {Emergent world representations: Exploring a sequence model trained on a synthetic task}.
\newblock In \emph{The Eleventh International Conference on Learning Representations}.

\bibitem[{Li et~al.(2023{\natexlab{b}})Li, Bubeck, Eldan, Del~Giorno, Gunasekar, and Lee}]{li2023textbooks}
Yuanzhi Li, S{\'e}bastien Bubeck, Ronen Eldan, Allie Del~Giorno, Suriya Gunasekar, and Yin~Tat Lee. 2023{\natexlab{b}}.
\newblock \href {https://arxiv.org/abs/2309.05463} {Textbooks are all you need ii: phi-1.5 technical report}.
\newblock \emph{arXiv preprint arXiv:2309.05463}.

\bibitem[{Liu et~al.(2019)Liu, Ott, Goyal, Du, Joshi, Chen, Levy, Lewis, Zettlemoyer, and Stoyanov}]{liu2019roberta}
Yinhan Liu, Myle Ott, Naman Goyal, Jingfei Du, Mandar Joshi, Danqi Chen, Omer Levy, Mike Lewis, Luke Zettlemoyer, and Veselin Stoyanov. 2019.
\newblock \href {https://arxiv.org/abs/1907.11692} {Roberta: A robustly optimized bert pretraining approach}.
\newblock \emph{arXiv preprint arXiv:1907.11692}.

\bibitem[{Lovering and Pavlick(2022)}]{lovering-pavlick-2022-unit}
Charles Lovering and Ellie Pavlick. 2022.
\newblock \href {https://doi.org/10.1162/tacl_a_00514} {Unit testing for concepts in neural networks}.
\newblock \emph{Transactions of the Association for Computational Linguistics}, 10:1193--1208.

\bibitem[{Marjieh et~al.(2022)Marjieh, Sucholutsky, Sumers, Jacoby, and Griffiths}]{marjieh2022predicting}
Raja Marjieh, Ilia Sucholutsky, Ted Sumers, Nori Jacoby, and Tom Griffiths. 2022.
\newblock \href {https://escholarship.org/uc/item/7mz2c7k4} {Predicting human similarity judgments using large language models}.
\newblock In \emph{Proceedings of the Annual Meeting of the Cognitive Science Society}, volume~44.

\bibitem[{McCoy et~al.(2023)McCoy, Yao, Friedman, Hardy, and Griffiths}]{mccoy2023embers}
R~Thomas McCoy, Shunyu Yao, Dan Friedman, Matthew Hardy, and Thomas~L Griffiths. 2023.
\newblock \href {https://arxiv.org/abs/2309.13638} {Embers of autoregression: Understanding large language models through the problem they are trained to solve}.
\newblock \emph{arXiv preprint arXiv:2309.13638}.

\bibitem[{Mihaylov et~al.(2018)Mihaylov, Clark, Khot, and Sabharwal}]{mihaylov-etal-2018-suit}
Todor Mihaylov, Peter Clark, Tushar Khot, and Ashish Sabharwal. 2018.
\newblock \href {https://doi.org/10.18653/v1/D18-1260} {Can a suit of armor conduct electricity? a new dataset for open book question answering}.
\newblock In \emph{Proceedings of the 2018 Conference on Empirical Methods in Natural Language Processing}, pages 2381--2391, Brussels, Belgium. Association for Computational Linguistics.

\bibitem[{Min et~al.(2022)Min, Lyu, Holtzman, Artetxe, Lewis, Hajishirzi, and Zettlemoyer}]{min-etal-2022-rethinking}
Sewon Min, Xinxi Lyu, Ari Holtzman, Mikel Artetxe, Mike Lewis, Hannaneh Hajishirzi, and Luke Zettlemoyer. 2022.
\newblock \href {https://doi.org/10.18653/v1/2022.emnlp-main.759} {Rethinking the role of demonstrations: What makes in-context learning work?}
\newblock In \emph{Proceedings of the 2022 Conference on Empirical Methods in Natural Language Processing}, pages 11048--11064, Abu Dhabi, United Arab Emirates. Association for Computational Linguistics.

\bibitem[{Misra et~al.(2022)Misra, Rayz, and Ettinger}]{misra2022property}
Kanishka Misra, Julia Rayz, and Allyson Ettinger. 2022.
\newblock \href {https://escholarship.org/uc/item/6170h6nj} {A property induction framework for neural language models}.
\newblock In \emph{Proceedings of the Annual Meeting of the Cognitive Science Society}, volume~44.

\bibitem[{Misra et~al.(2023)Misra, Rayz, and Ettinger}]{misra-etal-2023-comps}
Kanishka Misra, Julia Rayz, and Allyson Ettinger. 2023.
\newblock \href {https://doi.org/10.18653/v1/2023.eacl-main.213} {{COMPS}: Conceptual minimal pair sentences for testing robust property knowledge and its inheritance in pre-trained language models}.
\newblock In \emph{Proceedings of the 17th Conference of the European Chapter of the Association for Computational Linguistics}, pages 2928--2949, Dubrovnik, Croatia. Association for Computational Linguistics.

\bibitem[{Mitchell and Krakauer(2023)}]{mitchell_debate_2023}
Melanie Mitchell and David~C. Krakauer. 2023.
\newblock \href {https://doi.org/10.1073/pnas.2215907120} {The debate over understanding in ai’s large language models}.
\newblock \emph{Proceedings of the National Academy of Sciences}, 120(13):e2215907120.

\bibitem[{Murphy(2004)}]{murphy2004big}
Gregory Murphy. 2004.
\newblock \emph{The big book of concepts}.
\newblock MIT press.

\bibitem[{Partee et~al.(1984)}]{partee1984compositionality}
Barbara Partee et~al. 1984.
\newblock Compositionality.
\newblock \emph{Varieties of formal semantics}, 3:281--311.

\bibitem[{Patel and Pavlick(2022)}]{patel2022mapping}
Roma Patel and Ellie Pavlick. 2022.
\newblock \href {https://openreview.net/forum?id=gJcEM8sxHK} {Mapping language models to grounded conceptual spaces}.
\newblock In \emph{International Conference on Learning Representations}.

\bibitem[{Penedo et~al.(2023)Penedo, Malartic, Hesslow, Cojocaru, Cappelli, Alobeidli, Pannier, Almazrouei, and Launay}]{penedo2023refinedweb}
Guilherme Penedo, Quentin Malartic, Daniel Hesslow, Ruxandra Cojocaru, Alessandro Cappelli, Hamza Alobeidli, Baptiste Pannier, Ebtesam Almazrouei, and Julien Launay. 2023.
\newblock \href {https://arxiv.org/abs/2306.01116} {The refinedweb dataset for falcon llm: outperforming curated corpora with web data, and web data only}.
\newblock \emph{arXiv preprint arXiv:2306.01116}.

\bibitem[{Piantadosi and Hill(2022)}]{piantadosi2022meaning}
Steven Piantadosi and Felix Hill. 2022.
\newblock \href {https://openreview.net/forum?id=nRkJEwmZnM} {Meaning without reference in large language models}.
\newblock In \emph{NeurIPS 2022 Workshop on Neuro Causal and Symbolic AI (nCSI)}.

\bibitem[{Sap et~al.(2019)Sap, Rashkin, Chen, Le~Bras, and Choi}]{sap-etal-2019-social}
Maarten Sap, Hannah Rashkin, Derek Chen, Ronan Le~Bras, and Yejin Choi. 2019.
\newblock \href {https://doi.org/10.18653/v1/D19-1454} {Social {IQ}a: Commonsense reasoning about social interactions}.
\newblock In \emph{Proceedings of the 2019 Conference on Empirical Methods in Natural Language Processing and the 9th International Joint Conference on Natural Language Processing (EMNLP-IJCNLP)}, pages 4463--4473, Hong Kong, China. Association for Computational Linguistics.

\bibitem[{Siddique and Sufyan~Beg(2023)}]{siddique_reverse_2023}
Bushra Siddique and M.~M. Sufyan~Beg. 2023.
\newblock \href {https://doi.org/10.1007/s42979-022-01495-1} {Reverse {Dictionary} {Formation}: {State} of the {Art} and {Future} {Directions}}.
\newblock \emph{SN Computer Science}, 4(2):168.

\bibitem[{Speer(2022)}]{robyn_speer_2022_7199437}
Robyn Speer. 2022.
\newblock \href {https://doi.org/10.5281/zenodo.7199437} {rspeer/wordfreq: v3.0}.

\bibitem[{Suresh et~al.(2023)Suresh, Mukherjee, Yu, Huang, Padua, and Rogers}]{suresh-etal-2023-conceptual}
Siddharth Suresh, Kushin Mukherjee, Xizheng Yu, Wei-Chun Huang, Lisa Padua, and Timothy Rogers. 2023.
\newblock \href {https://doi.org/10.18653/v1/2023.emnlp-main.47} {Conceptual structure coheres in human cognition but not in large language models}.
\newblock In \emph{Proceedings of the 2023 Conference on Empirical Methods in Natural Language Processing}, pages 722--738, Singapore. Association for Computational Linguistics.

\bibitem[{Talmor et~al.(2019)Talmor, Herzig, Lourie, and Berant}]{talmor-etal-2019-commonsenseqa}
Alon Talmor, Jonathan Herzig, Nicholas Lourie, and Jonathan Berant. 2019.
\newblock \href {https://doi.org/10.18653/v1/N19-1421} {{C}ommonsense{QA}: A question answering challenge targeting commonsense knowledge}.
\newblock In \emph{Proceedings of the 2019 Conference of the North {A}merican Chapter of the Association for Computational Linguistics: Human Language Technologies, Volume 1 (Long and Short Papers)}, pages 4149--4158, Minneapolis, Minnesota. Association for Computational Linguistics.

\bibitem[{Team(2023)}]{MosaicML2023Introducing}
MosaicML~NLP Team. 2023.
\newblock \href {https://www.databricks.com/blog/mpt-7b} {Introducing mpt-7b: A new standard for open-source, commercially usable llms}.

\bibitem[{Touvron et~al.(2023{\natexlab{a}})Touvron, Lavril, Izacard, Martinet, Lachaux, Lacroix, Rozi{\`e}re, Goyal, Hambro, Azhar et~al.}]{touvron2023llama}
Hugo Touvron, Thibaut Lavril, Gautier Izacard, Xavier Martinet, Marie-Anne Lachaux, Timoth{\'e}e Lacroix, Baptiste Rozi{\`e}re, Naman Goyal, Eric Hambro, Faisal Azhar, et~al. 2023{\natexlab{a}}.
\newblock \href {https://arxiv.org/abs/2302.13971} {Llama: Open and efficient foundation language models}.
\newblock \emph{arXiv preprint arXiv:2302.13971}.

\bibitem[{Touvron et~al.(2023{\natexlab{b}})Touvron, Martin, Stone, Albert, Almahairi, Babaei, Bashlykov, Batra, Bhargava, Bhosale et~al.}]{touvron2023llama2}
Hugo Touvron, Louis Martin, Kevin Stone, Peter Albert, Amjad Almahairi, Yasmine Babaei, Nikolay Bashlykov, Soumya Batra, Prajjwal Bhargava, Shruti Bhosale, et~al. 2023{\natexlab{b}}.
\newblock \href {https://arxiv.org/abs/2307.09288} {Llama 2: Open foundation and fine-tuned chat models}.
\newblock \emph{arXiv preprint arXiv:2307.09288}.

\bibitem[{Vaswani et~al.(2017)Vaswani, Shazeer, Parmar, Uszkoreit, Jones, Gomez, Kaiser, and Polosukhin}]{vaswani_attention_2017}
Ashish Vaswani, Noam Shazeer, Niki Parmar, Jakob Uszkoreit, Llion Jones, Aidan~N Gomez, \L~ukasz Kaiser, and Illia Polosukhin. 2017.
\newblock \href {https://proceedings.neurips.cc/paper_files/paper/2017/file/3f5ee243547dee91fbd053c1c4a845aa-Paper.pdf} {Attention is all you need}.
\newblock In \emph{Advances in Neural Information Processing Systems}, volume~30. Curran Associates, Inc.

\bibitem[{Warstadt et~al.(2020)Warstadt, Parrish, Liu, Mohananey, Peng, Wang, and Bowman}]{warstadt-etal-2020-blimp-benchmark}
Alex Warstadt, Alicia Parrish, Haokun Liu, Anhad Mohananey, Wei Peng, Sheng-Fu Wang, and Samuel~R. Bowman. 2020.
\newblock \href {https://doi.org/10.1162/tacl_a_00321} {{BL}i{MP}: The benchmark of linguistic minimal pairs for {E}nglish}.
\newblock \emph{Transactions of the Association for Computational Linguistics}, 8:377--392.

\bibitem[{Webb et~al.(2023)Webb, Holyoak, and Lu}]{webb_emergent_2023}
Taylor Webb, Keith~J. Holyoak, and Hongjing Lu. 2023.
\newblock \href {https://doi.org/10.1038/s41562-023-01659-w} {Emergent analogical reasoning in large language models}.
\newblock \emph{Nature Human Behaviour}, 7(9):1526--1541.

\bibitem[{Wei et~al.(2022)Wei, Tay, Bommasani, Raffel, Zoph, Borgeaud, Yogatama, Bosma, Zhou, Metzler, Chi, Hashimoto, Vinyals, Liang, Dean, and Fedus}]{wei2022emergent}
Jason Wei, Yi~Tay, Rishi Bommasani, Colin Raffel, Barret Zoph, Sebastian Borgeaud, Dani Yogatama, Maarten Bosma, Denny Zhou, Donald Metzler, Ed~H. Chi, Tatsunori Hashimoto, Oriol Vinyals, Percy Liang, Jeff Dean, and William Fedus. 2022.
\newblock \href {https://openreview.net/forum?id=yzkSU5zdwD} {Emergent abilities of large language models}.
\newblock \emph{Transactions on Machine Learning Research}.

\bibitem[{Wolf et~al.(2019)Wolf, Debut, Sanh, Chaumond, Delangue, Moi, Cistac, Rault, Louf, Funtowicz et~al.}]{wolf2019huggingface}
Thomas Wolf, Lysandre Debut, Victor Sanh, Julien Chaumond, Clement Delangue, Anthony Moi, Pierric Cistac, Tim Rault, R{\'e}mi Louf, Morgan Funtowicz, et~al. 2019.
\newblock \href {https://arxiv.org/abs/1910.03771} {Huggingface's transformers: State-of-the-art natural language processing}.
\newblock \emph{arXiv preprint arXiv:1910.03771}.

\bibitem[{Wu et~al.(2023)Wu, Qiu, Ross, Aky{\"u}rek, Chen, Wang, Kim, Andreas, and Kim}]{wu2023reasoning}
Zhaofeng Wu, Linlu Qiu, Alexis Ross, Ekin Aky{\"u}rek, Boyuan Chen, Bailin Wang, Najoung Kim, Jacob Andreas, and Yoon Kim. 2023.
\newblock \href {https://arxiv.org/abs/2307.02477} {Reasoning or reciting? exploring the capabilities and limitations of language models through counterfactual tasks}.
\newblock \emph{arXiv preprint arXiv:2307.02477}.

\bibitem[{Yamakoshi et~al.(2023)Yamakoshi, McClelland, Goldberg, and Hawkins}]{yamakoshi-etal-2023-causal}
Takateru Yamakoshi, James McClelland, Adele Goldberg, and Robert Hawkins. 2023.
\newblock \href {https://doi.org/10.18653/v1/2023.findings-acl.839} {Causal interventions expose implicit situation models for commonsense language understanding}.
\newblock In \emph{Findings of the Association for Computational Linguistics: ACL 2023}, pages 13265--13293, Toronto, Canada. Association for Computational Linguistics.

\bibitem[{Yan et~al.(2020)Yan, Li, Qiu, and Deng}]{yan-etal-2020-bert}
Hang Yan, Xiaonan Li, Xipeng Qiu, and Bocao Deng. 2020.
\newblock \href {https://doi.org/10.18653/v1/2020.findings-emnlp.388} {{BERT} for monolingual and cross-lingual reverse dictionary}.
\newblock In \emph{Findings of the Association for Computational Linguistics: EMNLP 2020}, pages 4329--4338, Online. Association for Computational Linguistics.

\bibitem[{Zellers et~al.(2019)Zellers, Holtzman, Bisk, Farhadi, and Choi}]{zellers-etal-2019-hellaswag}
Rowan Zellers, Ari Holtzman, Yonatan Bisk, Ali Farhadi, and Yejin Choi. 2019.
\newblock \href {https://doi.org/10.18653/v1/P19-1472} {{H}ella{S}wag: Can a machine really finish your sentence?}
\newblock In \emph{Proceedings of the 57th Annual Meeting of the Association for Computational Linguistics}, pages 4791--4800, Florence, Italy. Association for Computational Linguistics.

\bibitem[{Zhang et~al.(2020)Zhang, Qi, Liu, Wang, Liu, and Sun}]{zhang2020multi}
Lei Zhang, Fanchao Qi, Zhiyuan Liu, Yasheng Wang, Qun Liu, and Maosong Sun. 2020.
\newblock \href {https://doi.org/10.1609/aaai.v34i01.5365} {Multi-channel reverse dictionary model}.
\newblock \emph{Proceedings of the AAAI Conference on Artificial Intelligence}, 34(01):312--319.

\bibitem[{Zheng et~al.(2019)Zheng, Pereira, Baker, and Hebart}]{zheng2018revealing}
Charles~Y. Zheng, Francisco Pereira, Chris~I. Baker, and Martin~N. Hebart. 2019.
\newblock \href {https://openreview.net/forum?id=ryxSrhC9KX} {Revealing interpretable object representations from human behavior}.
\newblock In \emph{International Conference on Learning Representations}.

\end{thebibliography}



\appendix

\section{Additional Materials for Reverse Dictionary as a Probe for Conceptual Inference}
\label{sec:appendix-find-concept}

\subsection{Comparison with Baselines}
\label{sec:appendix-find-concept-baseline}

Table~\ref{tab:concept-inference-baseline} compares the performance of LLMs with the baselines outlined in Section~\ref{sec:conceptual-probe}. Larger models generally achieve better performance, whereas they tend to be susceptible to noise introduced by demonstrations. However, the Pythia models (\texttt{pythia-1b4}, \texttt{pythia-6b9}, and \texttt{pythia-12b}) and \texttt{falcon-rw-7b} appear less sensitive to demonstrations, showing performance improvement over \textsc{NL} even when the pairings between descriptions and words are permuted, similar to previous research suggesting that some models may not heavily rely on the ground truth input-label mapping provided in the demonstrations \citep{min-etal-2022-rethinking}. Exploration of the phenomenon is left for future work. 

\begin{table}[!b]
\centering
\begin{tabular}{lcccc}
\noalign{\hrule height 1.2pt}
Model & \textsc{Demo} & \textsc{NL} & \textsc{Mis} & \textsc{Rand} \\
\noalign{\hrule height 0.75pt}
\texttt{pythia-1b4} & 46.5 & 16.2 & 35.0 & 24.3 \\
\texttt{pythia-2b8} & 52.4 & 25.9 & 5.5 & 6.1 \\
\texttt{pythia-6b9} & 60.1  & 30.6 & 47.0 & 52.5 \\
\texttt{pythia-12b} & 63.8 & 31.1 & 46.3 & 33.8 \\
\texttt{phi-1.5} & 52.1 & 28.1 & 6.6 & 26.3 \\
\texttt{phi-2} & 65.5 & 40.8 & 0.1 & 0.2 \\
\texttt{falcon-rw-1b} & 51.9  & 29.1 & 24.4 & 24.5 \\
\texttt{falcon-rw-7b} & 67.8 & 45.6 & 54.5 & 40.9 \\
\texttt{falcon-7b} & 72.5 & 39.5 & 1.7 & 4.5 \\
\texttt{mpt-7b} & 70.9 & 50.5 & 0.1 & 0.1 \\
\texttt{llama-7b} & 70.9 & 47.3 & 4.4 & 18.6 \\
\texttt{llama-13b} & 73.8 & 50.0 & 0.5 & 0.1 \\
\texttt{llama2-7b} & 73.0 & 49.5 & 1.0 & 0.4 \\
\texttt{llama2-13b} & 78.3 & 57.2 & 0.1 & 0.1 \\
\texttt{mistral-7b} & 77.6 & 58.0 & 1.8 & 0.1  \\
\noalign{\hrule height 1.2pt}
\end{tabular}
\caption{\label{tab:concept-inference-baseline}
Comparison of LLMs' performance (\textsc{Demo}) and the baselines with 24 demonstrations provided, except for \textsc{NL}, where the template is formatted in natural language with no demonstration.}
\end{table}

\subsection{Relationship between Conceptual Inference Ability and Model Size}

\label{sec:appendix-find-concept-corr-size}

\begin{figure}[t]
\centering
  \includegraphics[width=\linewidth]{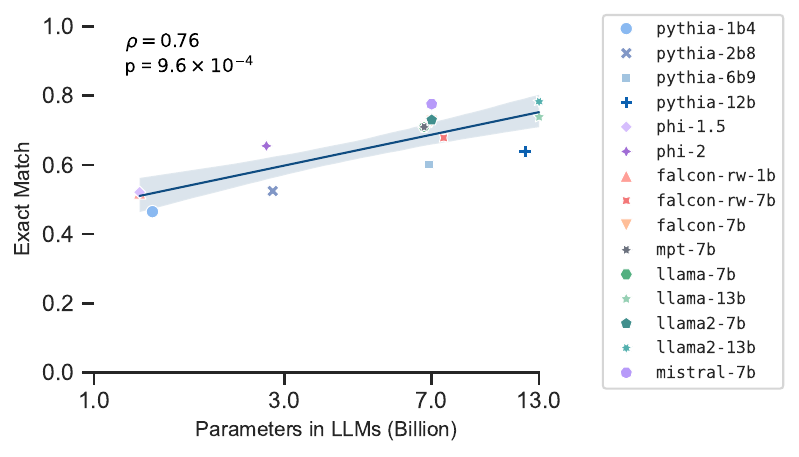}
  \caption{Correlation between the size of LLMs and their average conceptual inference ability measured as exact match accuracy on the reverse dictionary task  with 24 demonstrations provided.} \label{fig:corr-size}
\end{figure}

Figure~\ref{fig:corr-size} shows the relationship between the size of LLMs and their average performance in the reverse dictionary task when provided with 24 demonstrations. We notice a significant correlation.

\subsection{Impact of Variation in Descriptions}

\label{sec:appendix-desc-impact}

\paragraph{Setup}

As in Section~\ref{sec:conceptual-probe}, we randomly select 24 description--word pairs from THINGS as demonstrations and the query sentence is sourced from alternative databases: (1) 1,797 concepts in THINGS with descriptions obtained from WordNet\footnote{Out of the 1,854 concepts, 1,797 are linked with WordNet in THINGS.}, and (2) 200 pairs of words and human-written descriptions created by \citet{hill-etal-2016-learning-understand}, where the words are randomly chosen from the top 3000 most frequent tokens in the British National Corpus \citep{leech-etal-1994-claws4} but not within the top 100. There is no information about the synonyms of the words in \citet{hill-etal-2016-learning-understand}, which may affect the performance to some extent. We therefore also calculate the exact match performance based on the words themselves in terms of THINGS and WordNet for comparison. Additionally, we examine the robustness of LLMs to degraded syntactic structure by introducing varying degrees of word order permutations to the query description. Specifically, we take 30\%, 60\% and 100\% words from the query description in the THINGS database, randomly shuffle their order, and put them back to the description. For all our experiments here, we compute a model's average performance across 5 runs.

\begin{table}[t]
\centering
\begin{tabular}{lc}
\noalign{\hrule height 1.2pt}
Model & Hill200 \\
\noalign{\hrule height 0.75pt}
\texttt{pythia-1b4} & 41.8 \\
\texttt{pythia-6b9} & 48.7 \\
\texttt{falcon-rw-7b} & 62.4 \\
\texttt{falcon-7b} & 57.6 \\
\texttt{llama2-7b} & 67.3 \\
\texttt{llama2-13b} & 73.6 \\
\citet{zhang2020multi} & 32.0 \\
\citet{yan-etal-2020-bert} & 43.0 \\
\noalign{\hrule height 1.2pt}
\end{tabular}
\caption{\label{tab:concept-inference-baseline-related-papers}
Comparison of LLMs' performance with 24 demonstrations (\textsc{Demo}) and previous works \citep{zhang2020multi, yan-etal-2020-bert} on the Hill200 dataset. We use the reported accuracy@1 for comparison.}
\end{table}

\begin{figure*}[t]
\centering
  \includegraphics[width=.75\textwidth]{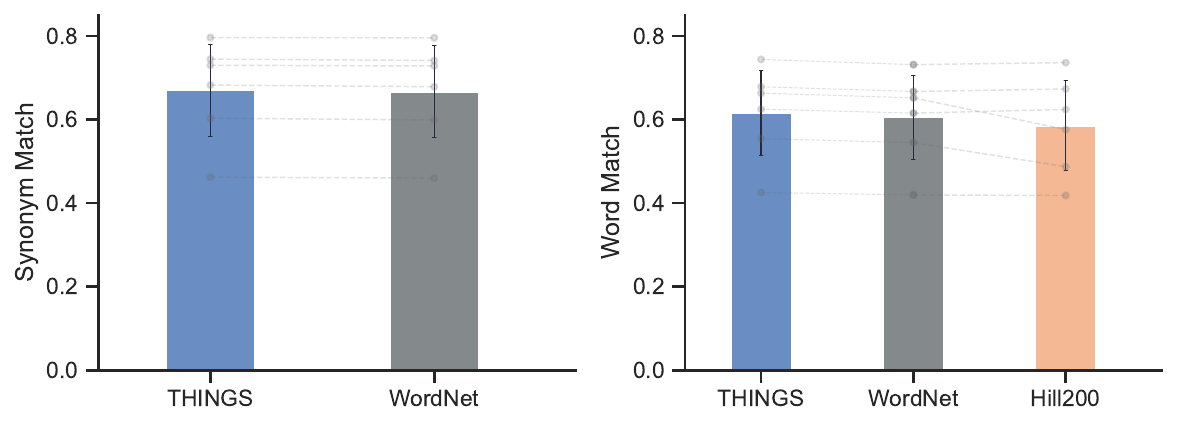}
  \caption{Performance of LLMs when confronted with various descriptions evaluated by exact matching of words or their synonyms. Larger models robustly adapts to diverse descriptions, and their performance is affected by the increasing degree of word order violations in the descriptions. Error bars represent standard error computed from the average performance of different models across 5 runs.} \label{fig:impact-desc}
\end{figure*}

\begin{figure*}[t]
\centering
  \includegraphics[width=.75\textwidth]{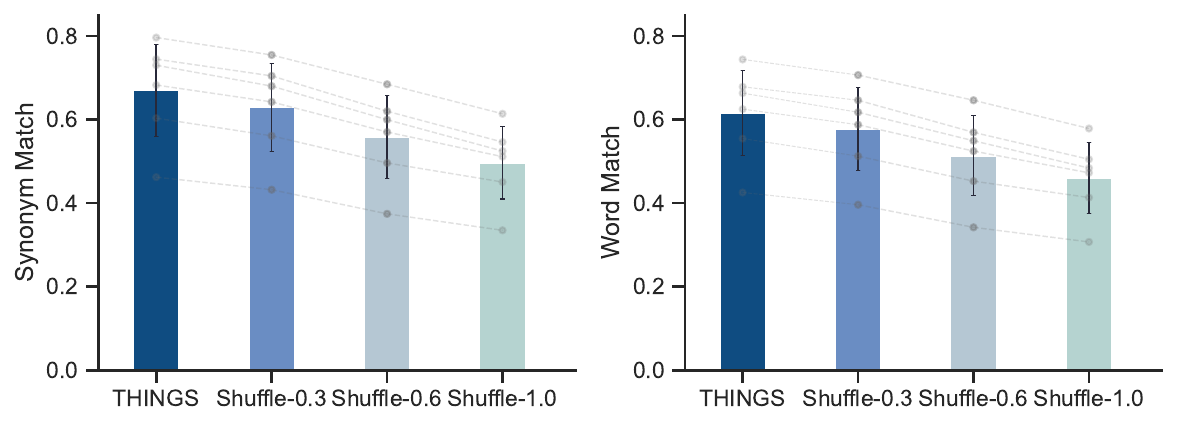}
  \caption{Performance of LLMs in the reverse-dictionary task when presented with descriptions in THINGS with varying degree of word order violations, evaluated by exact matching of words or their synonyms. Error bars represent standard error computed from the average performance of different models across 5 runs.} \label{fig:impact-shuffle}
\end{figure*}

\begin{figure*}[t]
\centering
  \includegraphics[width=\linewidth]{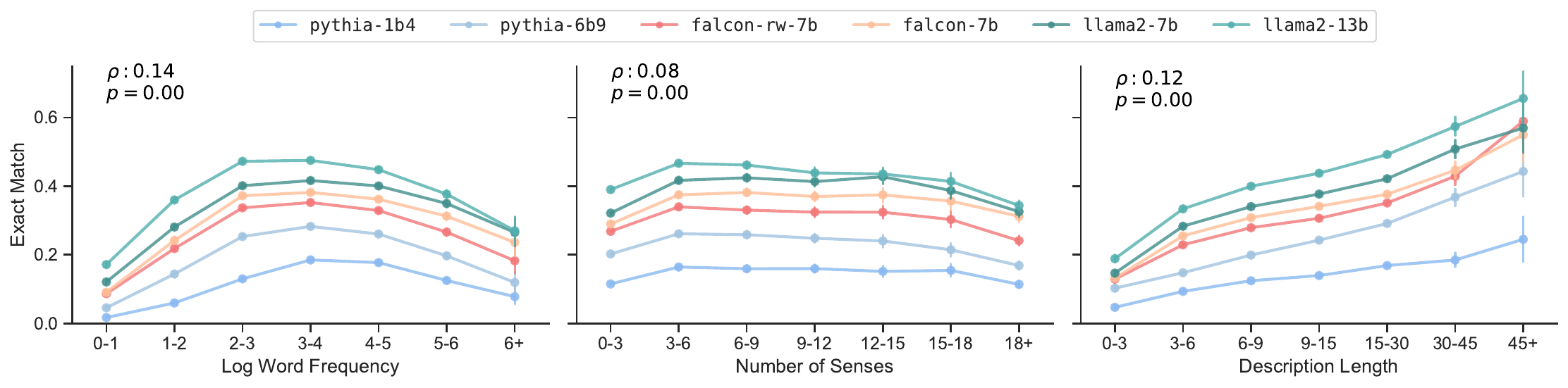}
  \caption{Impact of word frequency, number of senses and description length on the performance of LLMs in inferring concepts based on their descriptions. The log frequency of a word is calculated as the base-10 logarithm of its occurrence per billion words. The Spearman's correlation is averaged across different LLMs.} \label{fig:robust-query}
\end{figure*}

\begin{table*}[t]
\centering
\begin{tabular}{lcccc}
\noalign{\hrule height 1.2pt}
Model & Accuracy & WordFreq & NumSenses & DescLength \\
\noalign{\hrule height 0.75pt}
\texttt{pythia-1b4} & 12.8 & 0.148 & 0.068 & 0.088  \\
\texttt{pythia-6b9} & 21.7 & 0.136 & 0.061 &  0.138  \\
\texttt{falcon-rw-7b} & 28.6 & 0.131 & 0.070  &  0.114  \\
\texttt{falcon-7b} & 31.5 & 0.144 & 0.098 &  0.116  \\
\texttt{llama2-7b} & 34.9 & 0.144 & 0.102 & 0.127  \\
\texttt{llama2-13b} & 40.8 & 0.121 & 0.069 &  0.137 \\
\noalign{\hrule height 1.2pt}
\end{tabular}
\caption{\label{tab:wordnet-infl}
LLMs' performance in conceptual inference over the 117,659 words in WordNet, measured by exact match (Accuracy). The columns WordFreq, NumSenses, and DescLength represent the Spearman's rank correlation coefficients between accuracy and each of these three factors.}
\end{table*}

\paragraph{Results}

As shown in Figure~\ref{fig:impact-desc}, LLMs consistently maintain high performance across various descriptions, outperforming previous work explicitly training models including RoBERTa \citep{liu2019roberta} for the same task in Hill200 (Table~\ref{tab:concept-inference-baseline-related-papers}). We also note that the observed decline in performance for Hill200 may be attributable to the lack of synonym information. We observe modest effects of degraded syntactic structure on LLMs' performance on the reverse dictionary task, with degradation in performance becoming more pronounced as a higher degree of word order permutation is introduced (Figure~\ref{fig:impact-shuffle}). This shows some degree of robustness to input noise in LLMs and suggests that these models are at least sensitive to syntactic structure in the input when constructing conceptual representations.

\subsection{Impact of Query Properties}
\label{sec:appendix-query-impact}

\paragraph{Setup}

We randomly select 24 demonstrations from the THINGS database and test the performance of LLMs across the entire WordNet with 117,659 words in total. Due to the ambiguity of the pretraining corpus of LLMs, we use word frequencies from \citet{robyn_speer_2022_7199437} as a proxy, which is based on multiple sources such as Wikipedia and Books. The number of senses is directly obtained from WordNet, and the description length is determined by the word count of each description.

\paragraph{Results}

The performance of the models, along with the correlation between the performance and word frequency, number of senses, and description length, is illustrated in Table~\ref{tab:wordnet-infl} and Figure~\ref{fig:robust-query}.  Predicting words at the extremes of frequency proves challenging, akin to previous task-specific neural models that were explicitly trained for the reverse dictionary problem \citep{zhang2020multi, yan-etal-2020-bert}. The infrequent words can be more difficult for LLMs to learn, as suggested by previous work \citep{mccoy2023embers, chang-bergen-2022-word, pmlr-v202-kandpal23a}. Conversely, the most frequent words, such as \textit{be, have, do, make, take, use} etc., tend to be more polysemous \citep{CASAS201919} and may be inherently harder to describe precisely, which make them challenging to predict. The length of the description positively correlates the performance as well, possibly due to the provision of more comprehensive information in lengthier descriptions, facilitating the identification of the exact word.

\section{Additional Materials for the Analysis of Model Representations}
\label{sec:appendix-probing}

\subsection{Categorization}
\label{sec:appendix-probing-structure}

\begin{table*}
\centering
\begin{tabular}{lcccccc}
\noalign{\hrule height 1.2pt}
Model & \textsc{Demo} & \textsc{NL} & \textsc{Mis} & \textsc{W2W} & \textsc{Word} & \textsc{Descr} \\
\noalign{\hrule height 0.75pt}
\texttt{pythia-1b4} & 88.0  & 81.9 & 86.3 & 65.8 & 51.4 & 65.6 \\
\texttt{pythia-2b8} & 89.7 & 84.4 & 79.5 & 78.0 & 57.9 & 69.5 \\
\texttt{pythia-6b9} & 90.7 & 83.2 & 89.5 & 84.4 & 59.9 & 72.6 \\
\texttt{pythia-12b} & 90.7  & 82.4 & 88.3 & 84.7 & 59.6 & 74.4 \\
\texttt{phi-1.5} & 89.2 & 81.3 & 82.3 & 80.0 & 60.4 & 72.1 \\
\texttt{phi-2} & 91.4  & 85.6 & 39.4 & 84.5 & 70.5 & 66.7 \\
\texttt{falcon-rw-1b} & 89.1 & 87.7 & 84.3 & 81.1 & 66.6 & 74.4 \\
\texttt{falcon-rw-7b} & 90.4 & 87.7 & 90.5 & 86.2 & 55.9 & 75.1 \\
\texttt{falcon-7b} & 90.6 & 79.6 & 73.5 & 78.0 & 31.5 & 56.8 \\
\texttt{mpt-7b} & 90.3 & 89.0 & 61.1 & 81.9 & 39.8 & 75.5 \\
\texttt{llama-7b} & 90.6 & 54.0 & 63.8 & 71.5 & 68.4 & 58.4 \\
\texttt{llama-13b} & 89.5 & 54.3 & 57.6 & 38.0 & 62.3 & 62.3 \\
\texttt{llama2-7b} & 89.0 & 71.1 & 72.8 &  44.0 & 60.9 & 67.6 \\
\texttt{llama2-13b} & 90.4  & 86.2 & 57.6 & 87.1 & 70.1 & 75.9 \\
\texttt{mistral-7b} & 91.5 & 87.4 & 45.0 & 86.7 & 60.7 & 73.7 \\
\noalign{\hrule height 0.75pt}
\textsc{fastText} & \multicolumn{6}{c}{77.9} \\
\textsc{SPoSE} & \multicolumn{6}{c}{85.9} \\
\noalign{\hrule height 1.2pt}
\end{tabular}
\caption{\label{tab:categorization}
Accuracy of using representations derived from LLMs under the reverse dictionary task (\textsc{Demo}) and other baseline representations for similarity-based categorization. \textsc{Demo}, \textsc{Perm}, and \textsc{Mis} are representations derived from LLMs with 24 demonstrations provided. \textsc{Descr} denotes the \textsc{Description} baseline where we take the representations of LLMs prior to encountering the delimiter ``$\Rightarrow$''.}
\end{table*}

\paragraph{Method}

For categorization, we leave each concept out in turn and compute the centroid for each category by averaging the representations of the remaining concepts within it. The classification is based on the cosine distance between the concept and each centroid. 

\paragraph{Data}

Following \citet{hebart_revealing_2020}, we remove subcategories of other categories, concepts belonging to multiple categories and categories with fewer than ten concepts. This results in 18 out of 27 categories in THINGS, including animal, body part, clothing, container, electronic device, food, furniture, home decoration, medical equipment, musical instrument, office supply, part of car, plant, sports equipment, tool, toy, vehicle and weapon. These categories comprise 1,112 concepts.

\paragraph{Results}

Table~\ref{tab:categorization} presents the categorization results for all LLMs and baselines. LLMs generally achieve an average performance at around 90\% for THINGS, surpassing all the baselines including \textsc{fastText} and \textsc{SPoSE}. The \textsc{NL} baseline achieve a relatively high accuracy, in line with its performance in the reverse dictionary task.

\subsection{Feature Ratings}
\label{sec:appendix-probing-feature}

\paragraph{Data}

As described in Section~\ref{sec:representation}, we use the XCSLB feature norm \citep{misra2022property} for our analysis, which expands the original CSLB dataset \citep{devereux_centre_2014} with necessary modifications for consistency. XCSLB includes 3,645 descriptive features for 521 concepts. We take the concepts that overlap with those in THINGS and remove features that are too sparse with fewer than 20 concepts. This results in 257 features associated with 388 concepts in total.

\paragraph{Results}

\begin{figure*}[t]
\centering
  \includegraphics[width=\textwidth]{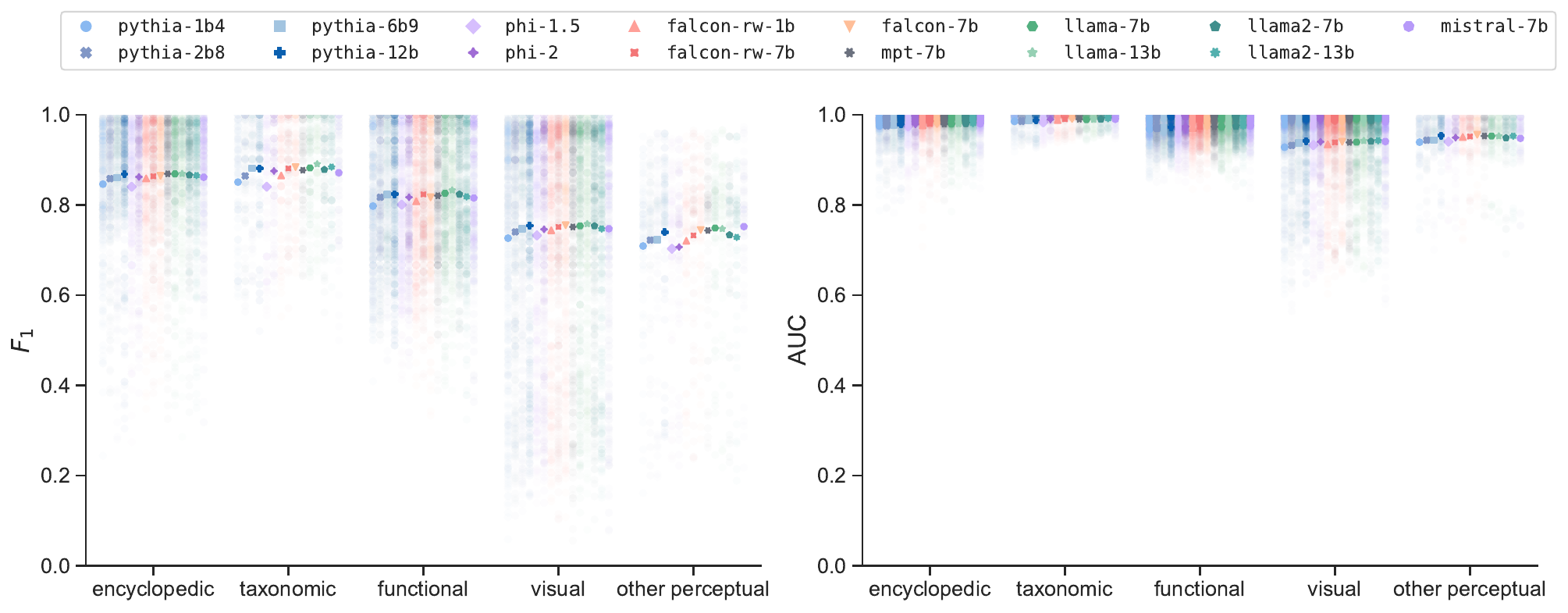}
  \caption{Performance of using LLMs' representations to predict the object features in XCSLB. Performance is measured by $F_{1}$ score (Left) and AUC (Right). Each point denotes a feature of a certain type.} \label{fig:xcslb-all}
\end{figure*}

The results for feature prediction of LLMs in XCSLB, measured by $F_{1}$ score and AUC, are depicted in Figure~\ref{fig:xcslb-all}. The comparison with baselines is presented in Table~\ref{tab:feature-prediction}.

\begin{table*}
\centering
\begin{tabular}{lcccccc}
\noalign{\hrule height 1.2pt}
Model & \textsc{Demo} & \textsc{NL} & \textsc{Mis} & \textsc{W2W} & \textsc{Word} & \textsc{Descr} \\
\noalign{\hrule height 0.75pt}
\texttt{pythia-1b4} & 78.6 / 95.7 & 75.6 / 95.4 & 76.0 / 95.3 & 66.6 / 93.7 & 63.6 / 90.5 & 66.5 / 93.1 \\
\texttt{pythia-2b8} & 80.1 / 95.9 & 77.5 / 95.7 & 74.3 / 94.9 & 74.6 / 95.6 & 65.5 / 91.7 & 69.2 / 94.1 \\
\texttt{pythia-6b9} & 80.6 / 96.1 & 77.7 / 95.7 & 79.3 / 95.8 & 77.9 / 96.5 & 68.4 / 92.6 & 69.9 / 94.4 \\
\texttt{pythia-12b} & 81.2 / 96.4 & 78.0 / 96.0 & 80.1 / 96.1 & 79.7 / 96.8 & 69.1 / 93.3 & 70.4 / 94.6 \\
\texttt{phi-1.5} & 78.6 / 95.8 & 75.8 / 95.3 & 74.2 / 94.8 & 75.5 / 95.5 & 67.6 / 92.1 & 67.7 / 93.6 \\
\texttt{phi-2} & 80.4 / 96.4 & 78.0 / 96.0 & 68.8 / 93.3 & 79.9 / 96.9 & 73.9 / 94.8 & 68.6 / 94.0  \\
\texttt{falcon-rw-1b} & 80.0 / 96.1 & 77.3 / 95.6 & 76.3 / 95.1 & 75.8 / 95.9 & 69.1 / 92.3  & 68.1 / 93.8 \\
\texttt{falcon-rw-7b} & 80.9 / 96.4 & 79.0 / 96.2 & 80.0 / 96.1 & 77.6 / 96.5 & 69.2 / 92.6 & 71.1 / 94.9 \\
\texttt{falcon-7b} & 81.0 / 96.5 & 79.2 / 96.2 & 75.2 / 94.7 & 77.2 / 95.8 & 71.2 / 92.8 & 67.9 / 93.4 \\
\texttt{mpt-7b} & 81.0 / 96.4 & 79.8 / 96.2 & 73.2 / 94.8 & 78.1 / 96.6 & 71.9 / 94.0 & 71.4 / 95.1 \\
\texttt{llama-7b} & 81.3 / 96.4 & 78.6 / 95.9 & 77.2 / 94.9 & 78.4 / 96.8 & 75.9 / 95.4 & 69.1 / 94.1 \\
\texttt{llama-13b} & 81.7 / 96.5 & 78.5 / 96.1 & 74.8 / 94.6 & 79.0 / 96.8 & 74.2 / 94.9 & 69.6 / 94.4 \\
\texttt{llama2-7b} & 81.1 / 96.5 & 79.8 / 96.2 & 75.3 / 95.0 & 77.2 / 96.3 & 72.9 / 94.6 & 70.1 / 94.6 \\
\texttt{llama2-13b} & 80.7 / 96.6 & 79.8 / 96.4 & 69.3 / 93.9 & 79.3 / 96.7 & 76.7 / 95.5 & 66.5 / 94.5 \\
\texttt{mistral-7b} & 80.6 / 96.5 & 79.7 / 96.3 & 74.3 / 94.6 & 79.4 / 96.8 & 75.8 / 95.3 & 69.8 / 94.7 \\
\noalign{\hrule height 0.75pt}
\textsc{fastText} & \multicolumn{6}{c}{76.3 / 95.1} \\
\textsc{SPoSE} & \multicolumn{6}{c}{68.4 / 92.4} \\
\noalign{\hrule height 1.2pt}
\end{tabular}
\caption{\label{tab:feature-prediction}
Performance of LLMs (\textsc{Demo}) and other baselines in predicting semantic features in XCSLB evaluated by the average $F_{1}$ (/AUC) score. \textsc{Demo} and \textsc{Mis} are the representations derived from LLMs with 24 demonstrations provided. \textsc{Descr} denotes the \textsc{Description} baseline where we take the representations of LLMs prior to encountering the delimiter.}
\end{table*}

\section{Additional Materials for Relationship between Conceptual Inference and General Abilities}

\label{sec:appendix-corr-general}

\subsection{Details of Evaluation}

\label{sec:appendix-corr-general-evaluation}

Considering the multiple-choice format of the reasoning tasks, let $w_{1:n}$ be the prompt composed of $n$ tokens, and $w_{n+1:c_{i}}$ denote the $i$-th possible answer with $c_{i} - n$ tokens among all candidates $\mathcal{C}$. We evaluate LLMs by their accuracy in ranking the correct answer with the highest probability, where the score of each answer is calculated as $\sum_{t=n+1}^{c_{i}}\log p_{\mathcal{M}}\left(w_{t} \mid w_{<t}\right)$. 

\subsection{Results}
\label{sec:appendix-corr-general-results}

\begin{figure*}[t]
\centering
  \includegraphics[width=\textwidth]{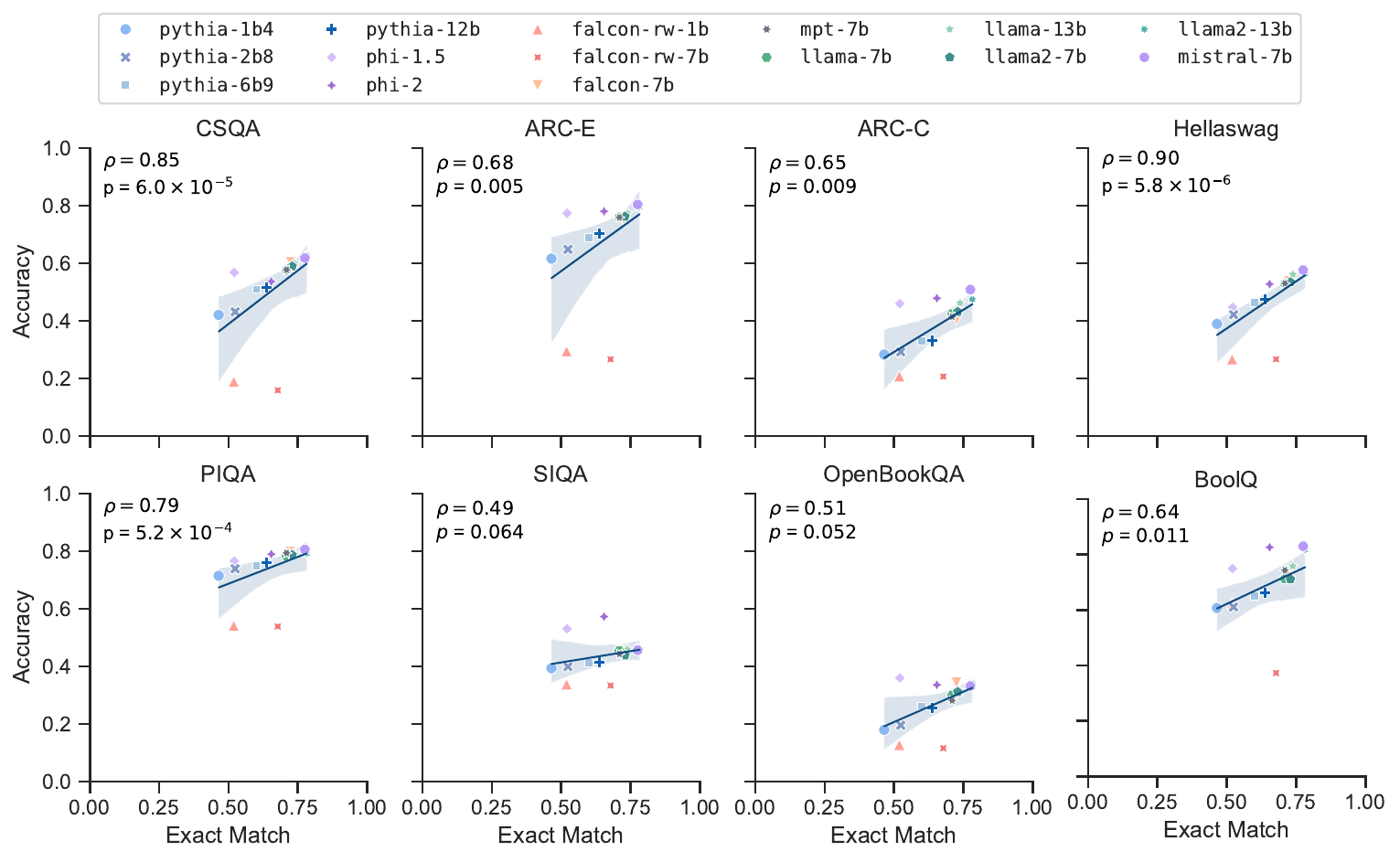}
  \caption{Correlation between LLMs' performance across different reasoning tasks and their average performance in conceptual inference with 24 demonstrations provided. The significant correlation across different tasks suggests a pivotal role of conceptual inference in LLMs' general ability.} \label{fig:corr-reasoning-all-tasks}
\end{figure*}

The correlation between LLMs' performance in conceptual inference and their performance in each reasoning task is shown in Figure~\ref{fig:corr-reasoning-all-tasks}.

\section{Additional Materials for Relationship between Conceptual Inference and Syntactic Generalization}
\label{sec:appendix-syntax}

\begin{figure*}[t]
\centering
  \includegraphics[width=\textwidth]{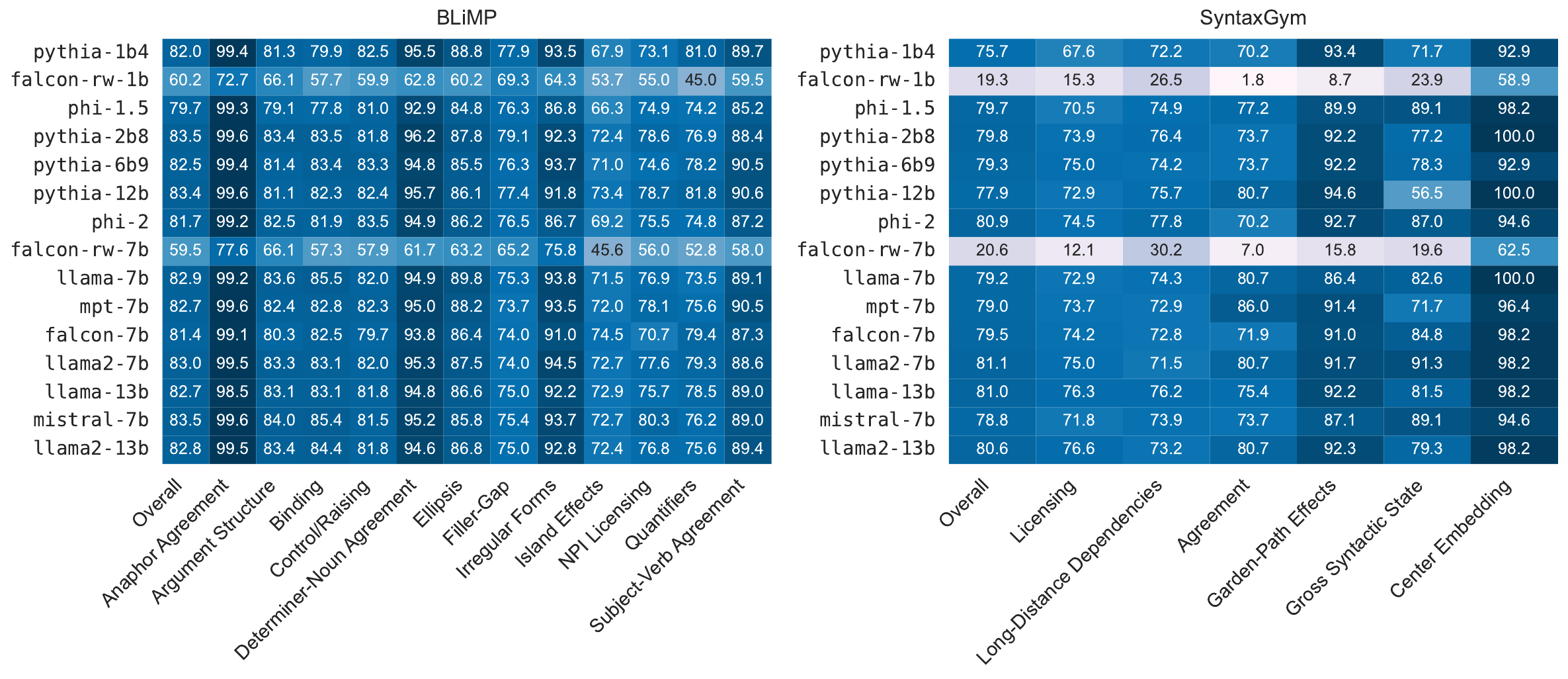}
  \caption{Performance of LLMs across different linguistic phenomena in BLiMP and SyntaxGym. The LLMs are ranked by their average performance in conceptual inference with 24 demonstrations.} \label{fig:syntax-details}
\end{figure*}

LLMs' performance across different linguistic phenomena tested in BLiMP and SyntaxGym are shown in Figure~\ref{fig:syntax-details}. The lack of correlation, along with the inferior performance of \texttt{falcon-rw} models, suggests that LLMs' syntactic generalization ability might be dissociable from their capacity to construct conceptual representations.

\section{Additional Materials for Generalizing Reverse Dictionary to Commonsense Reasoning}
\label{sec:appendix-generalize}

\subsection{Details of Setup}
\label{sec:appendix-generalize-evaluation}

The ground truth answers for ProtoQA consist of a ranked list of clusters of answers collected from humans. Similar to \citet{boratko-etal-2020-protoqa}, we use Nucleus Sampling \citep{Holtzman2020The} to get 100 sampled answers from LLMs per question, sort the answers by frequency counts, and obtain a ranked list of 10 answers ordered from most to least common. The answers are then matched with clusters of ground truth answers. In terms of exact match, the answers generated by LLMs are compared with those within each cluster, receiving a score of 1 if they match any string in it and 0 otherwise. For WordNet match, the answers generated by LLMs are tokenized and match with the synsets in WordNet associated with the gold answers. The overall score is computed based on a reward matrix where each cluster's size determines the reward assigned if the generated answers achieve a score of 1. For more details, see \citet{boratko-etal-2020-protoqa}. 

For this experiment, we select three LLMs across various model series that demonstrate relatively good performance in the reverse dictionary task, including \texttt{llama2-13b}, \texttt{llama2-7b} and \texttt{falcon-7b}. During generation, we set the max tokens to 28, and both \texttt{top\_p} and temperature to 1.0, as well as a repetition penalty of 1.0.

\subsection{Results}

\label{sec:appendix-generalize-results}

\paragraph{Impact of conceptual inference on ProtoQA}

The performance of LLMs in ProtoQA under different conditions is shown in Table~\ref{tab:protoqa-causal}.

\begin{table*}[t]
    \centering
    \begin{adjustbox}{width=\textwidth}
    \begin{tabular}{p{0.1\textwidth}l|cccc|ccc|cccc|ccc}
    \noalign{\hrule height 1.2pt}
    \multirow{3}{*}{} & \multirow{3}{*}{}  & \multicolumn{7}{c}{Exact Match} & \multicolumn{7}{c}{WordNet Match} \\ \cline{3-16}
    & & \multicolumn{4}{c}{Max Answers} & \multicolumn{3}{c}{Max Incorrect} & \multicolumn{4}{c}{Max Answers} & \multicolumn{3}{c}{Max Incorrect}  \\ \cline{3-16}
    & & 1 & 3 & 5 & 10 & 1 & 3 & 5 & 1 & 3 & 5 & 10 & 1 & 3 & 5 \\
    \noalign{\hrule height 0.75pt}
    $\textrm{Human}^{\ast}$ & & 78.4 & 74.4 & 72.5 & 73.3 & 55.8 & 69.4 & 72.4 & 78.4 & 76.8 & 76.0 & 77.0 & 59.0 & 74.0 & 77.9 \\
    \hline
    \multirow{1}{*}{$\textrm{GPT-2}^{\ast}$} & \textsc{NL} & 5.6 & 15.9 & 18.3 & 23.2 & 3.3 & 15.1 & 19.3 & 6.2 & 18.5 & 23.0 & 30.5 & 4.3 & 17.9 & 24.2 \\
    \hline
    \multirow{4}{*}{\parbox{0.1\textwidth}{Falcon 7B}} & \textsc{NL} & 17.4 & 15.2 & 16.0 & 15.2 & 8.2 & 13.3 & 14.5 & 24.6 & 25.8 & 27.5 & 27.9 & 13.0 & 21.4 & 24.7 \\
     & 1 & 18.4 & 21.5 & 20.7 & 20.9 & 10.4 & 17.9 & 19.5 & 19.1 & 24.0 & 23.6 & 26.8 & 12.2 & 19.9  & 22.1 \\
     & 12 & 21.0 & 21.9 & 23.4 & 27.9 & 12.1 & 19.9 & 22.7 & 22.5 & 25.1 & 27.3 & 31.5 & 13.3 & 23.9 & 26.5 \\
     & 24 & 21.3 & 23.6 & 25.1 & 29.5 & 13.0 & 21.7 & 24.5 & 23.1 & 27.5 & 30.5 & 34.2 & 14.8 & 25.1 & 30.7 \\
    \hline
    \multirow{4}{*}{\parbox{0.1\textwidth}{LLaMA2 7B}} & \textsc{NL} & 17.0 & 19.4 & 18.4 & 17.3 & 9.3 & 16.2 & 16.5 & 21.4 & 27.5 & 28.6 & 32.3 & 12.5 & 22.7 & 26.6 \\
     & 1 & 11.0 & 12.8 & 13.0 & 13.9 & 6.1 & 10.8 & 12.6 & 13.5 & 14.5 & 15.2 & 7.3 & 7.3 & 12.3 & 14.5 \\
     & 12  & 19.4 & 20.8 & 22.7 & 25.9 & 10.4 & 17.0 & 22.6 & 22.8 & 24.2 & 26.0 & 31.2 & 13.1 & 19.6 & 25.4 \\
     & 24  & 15.4 & 20.7 & 26.2 & 29.9 & 7.7 & 17.8 & 25.8 & 18.7 & 22.8 & 29.1 & 34.6 & 9.5 & 20.7 & 29.3 \\
     \hline
    \multirow{4}{*}{\parbox{0.1\textwidth}{LLaMA2 13B}} & \textsc{NL} & 19.1 & 19.2 & 17.7 & 16.3 & 11.5 & 15.6 & 16.0 & 25.8 & 26.1 & 25.8 & 25.9 & 14.8 & 21.7 & 23.6 \\
     & 1 & 16.0 & 20.9 & 21.2 & 24.3 & 7.3 & 17.6 & 20.6 & 19.0 & 24.4 &  26.5 & 29.7 & 9.3 & 22.0 & 25.5 \\
     & 12  & 19.9 & 20.4 & 22.6 & 26.8 & 11.0 & 18.5 & 23.4 & 22.7 & 23.8 & 26.4 & 31.9 & 13.7 & 22.5 & 27.6 \\
     & 24  & 22.0 & 23.8 & 26.6 & 31.1 & 12.8 & 21.6 & 25.6 & 25.4 & 28.3 & 33.3 & 37.6 & 15.1 & 26.9 & 32.6 \\
    \noalign{\hrule height 1.2pt}
    \end{tabular}
    \end{adjustbox}
\caption{Performance of humans and LLMs in ProtoQA. Data of $\textrm{GPT-2}^{\ast}$ and human performance are obtained from the original paper \citep{boratko-etal-2020-protoqa}.}
\label{tab:protoqa-causal}
\end{table*}

\paragraph{Examples of LLM-generated answers}

Table~\ref{tab:protoqa-llm-answers} illustrates several instances of LLM-generated answers for ProtoQA under different conditions.

\begin{table*}[t]
    \centering
    \begin{tabular}{lm{.75\textwidth}}
    \noalign{\hrule height 1.2pt}
    Question & Name a sport that requires a lot of equipment. \\
    \noalign{\hrule height .75pt}
    Human & football, hockey, soccer, baseball, golf, motocross, basketball, fencing, high jump, sky-diving, snowboard, swimming, tennis, ... \\
    \hline
    \textsc{NL} & ``SEBA inline hockey The other two are ice hockey and roller hockey What is an inline hockey stick Ice hockey          has,'' \par ``wrestling Wrestling mats used for judo judo and judo training are one of the most popular accessories,'' \par ``sailing         This popular sport can be extremely expensive if you arent careful One of the more common items that people will need is,'' \par ``basketball The league will produce 17 Tag games during the 2018 MLB season which will be held as part,'' \par ``golf Aside from       tickets to 18 holes  and maybe a sleeve of golf balls  golfers can get away,'' \par ``beach volleyball Players require shoes knee      pads shorts and at least,'' \par ``IHL hockey which requires ten or more people to play The other sports that need more than seven     people are boxing rodeo,'' \par ``rock climbing Head to an indoor rock climbing gym or check out a local gym that offers indoor        bouldering,'' ... \\
    \hline
    \textsc{Demo}24  & ``golf,'' ``gear,'' ``tennis,'' ``skiing,'' ``football,'' ``rugby,'' ``lawn bowling,'' ``ball game,'' ``fishing,'' ``cricket,'' ... \\
    \noalign{\hrule height 1.2pt}\\
    \noalign{\hrule height 1.2pt}
    Question & Name something that people usually do before they leave the house for work? \\
    \noalign{\hrule height .75pt}
    Human & brush their teeth, go to the bathroom, showering, wash, bath, eat something, have breakfast, drink coffee, dress up, grab keys, lock the door, turn the lights off, comb, look in the mirror, say goodbye, jogging, pray, ... \\
    \hline
    \textsc{NL} & ``to check the weather forecast read more,'' \par ``have a cup of coffee This beverage gives them the extra energy that       they need so they can complete their daily tasks and finish their work,'' \par ``have breakfast A healthy breakfast can ensure         people start the day feeling satisfied and energized You can then translate your satisfaction into the work,'' \par ``turn on the      radio It is a reflex And this was a Wall Street Journal poll People dont think about how they,'' \par ``to have breakfast Same         thing with your GSD It would be important that you provide proteinpacked food to your dog for its,'' \par ``to walk the dog And        when they come back they already are looking for their tired little friend to have a nice cuddle,'' \par ``shop Groceries              housewares home improvement stores etc For me I can only buy things that I can carry on,'' \par ``to have a early morning coffee       People who dont like coffee usually prefers tea But most people drink coffee everyday The,'' \par "take a smoke For chainsmokers      this is a daily ritual that could cost them an arm and a leg if they, '' ...\\
    \hline
    \textsc{Demo}24 & ``commute,'' ``breakfast,'' ``get dressed,'' ``brush teeth'', ``morning routine,'' ``get ready,'' ``wake up,'' ``shower,'' ``shave,'' ... \\
    \noalign{\hrule height 1.2pt}
    \end{tabular}
\caption{Some examples of the answers to the questions in ProtoQA generated by LLaMA2-13B under different conditions.}
\label{tab:protoqa-llm-answers}
\end{table*}

\section{Implementation Details}

\subsection{Large Language Models}

\label{sec:appendix-llms}

Detailed information about the LLMs used in our experiments is presented in Table~\ref{tab:llms-details}.

\begin{table*}
\centering
\begin{tabular}{llm{.25\textwidth}l}
\noalign{\hrule height 1.2pt}
\textbf{Series} & \textbf{Models} & \textbf{Dataset} & \textbf{\#Tokens}\\
\hline
Falcon & \makecell[l]{\href{https://huggingface.co/tiiuae/falcon-rw-1b}{\texttt{tiiuae/falcon-rw-1b}} \\ \href{https://huggingface.co/tiiuae/falcon-rw-7b}{\texttt{tiiuae/falcon-rw-7b}} \\ \href{https://huggingface.co/tiiuae/falcon-7b}{\texttt{tiiuae/falcon-7b}}} & RefinedWeb \par (enhanced with curated corpora like the Pile) & \makecell[l]{350B \\ 350B \\ 1.5T}   \\
\hline
LLaMA 1 & \makecell[l]{\href{https://huggingface.co/huggyllama/llama-7b}{\texttt{huggyllama/llama-7b}} \\ \href{https://huggingface.co/huggyllama/llama-13b}{\texttt{huggyllama/llama-13b}}} & CommonCrawl, C4, Github, Wikipedia, Books, ArXiv, StackExchange & 1T \\
\hline
LLaMA 2 & \makecell[l]{\href{https://huggingface.co/meta-llama/Llama-2-7b}{\texttt{meta-llama/Llama-2-7b}} \\ \href{https://huggingface.co/meta-llama/Llama-2-13b}{\texttt{meta-llama/Llama-2-13b}}} & data from publicly available sources & 2T \\
\hline
Mistral & \makecell[l]{\href{https://huggingface.co/mistralai/Mistral-7B-v0.1}{\texttt{mistralai/Mistral-7B-v0.1}}} & & \\
\hline
MPT & \makecell[l]{\href{https://huggingface.co/mosaicml/mpt-7b}{\texttt{mosaicml/mpt-7b}}} & mC4, C4, RedPajama, the Stack Dedup & 1T \\
\hline
Phi & \makecell[l]{\href{https://huggingface.co/microsoft/phi-1_5}{\texttt{microsoft/phi-1\_5}} (1.3b) \\ \href{https://huggingface.co/microsoft/phi-2}{\texttt{microsoft/phi-2}} (2.7b)} & code-language and synthetic data (augmented with filtered web data) & \makecell[l]{ 30B \\ 1.4T} \\
\hline
Pythia & \makecell[l]{\href{https://huggingface.co/EleutherAI/pythia-1.4b-deduped}{\texttt{EleutherAI/pythia-1.4b-deduped}} \\ \href{https://huggingface.co/EleutherAI/pythia-2.8b-deduped}{\texttt{EleutherAI/pythia-2.8b-deduped}} \\ \href{https://huggingface.co/EleutherAI/pythia-6.9b-deduped}{\texttt{EleutherAI/pythia-6.9b-deduped}} \\ \href{https://huggingface.co/EleutherAI/pythia-12b-deduped}{\texttt{EleutherAI/pythia-12b-deduped}}} & Pile (deduplicated) & 300B \\
\noalign{\hrule height 1.2pt}
\end{tabular}
\caption{LLMs used for our experiments. The dataset column for \texttt{mistral-7b} is empty due to lack of information about its pretraining data.}
\label{tab:llms-details}
\end{table*}

\subsection{Prompt Templates}
\label{sec:appendix-generalize-templates}

Table~\ref{tab:reasoning-templates} shows the prompt templates in terms of \textsc{NL} for all the reasoning tasks. The prompt templates for ProtoQA are presented in Table~\ref{tab:protoqa-templates}, as adopted from the original paper \citep{boratko-etal-2020-protoqa}.

\subsection{Hyperparameters}

We set the max tokens to 28 for all generation tasks. In terms of ProtoQA involving nucleus sampling, we set both \texttt{top\_p} and temperature to 1.0, alongside a repetition penalty of 1.0, to ensure a fair comparison across models.

\begin{table*}[t]
    \centering
    \begin{tabular}{p{.425\textwidth}p{.425\textwidth}}
    \noalign{\hrule height 1.2pt}
    Dataset & \textsc{NL} Template \\
    \noalign{\hrule height 0.75pt}
    CSQA & \makecell[l]{Question: [\texttt{Question}] \\ Answer: [\texttt{Answer}]} \\
    \hline
    ARC (E \& C) & \makecell[l]{Question: [\texttt{Question}] \\ Answer: [\texttt{Answer}]} \\
    \hline
    HellaSwag & \makecell[l]{Question: [\texttt{Question}] \\ Answer: [\texttt{Answer}]} \\
    \hline
    PIQA & \makecell[l]{Goal: [\texttt{Question}] \\ Answer: [\texttt{Answer}]} \\
    \hline
    SIQA & \makecell[l]{[\texttt{Context}] \\ Question: [\texttt{Question}] \\ Answer: [\texttt{Answer}]} \\
    \hline
    OpenbookQA & \makecell[l]{Question: [\texttt{Question}] \\ Answer: [\texttt{Answer}]} \\
    \hline
    BoolQ & \makecell[l]{[\texttt{Context}] \\ Question: [\texttt{Question}] \\ Answer: [\texttt{Answer}]} \\
    \noalign{\hrule height 1.2pt}
    \end{tabular}
\caption{Prompt templates for various reasoning tasks in \textsc{NL}.}
\label{tab:reasoning-templates}
\end{table*}

\begin{table*}[t]
    \centering
    \begin{tabular}{p{.425\textwidth}p{.425\textwidth}}
    \noalign{\hrule height 1.2pt}
    ProtoQA Question & \textsc{NL} Template \\
    \noalign{\hrule height 0.75pt}
    Name something ... [\texttt{Answer}] & One thing ... is [\texttt{Answer}] \\
    Tell me something ... [\texttt{Answer}] & One thing ... is [\texttt{Answer}] \\
    Name a(/an) ... [\texttt{Answer}] & One ... is [\texttt{Answer}] \\
    How can you tell ... [\texttt{Answer}] & One way to tell ... is [\texttt{Answer}] \\
    Give me a(/an) ... [\texttt{Answer}] & One ... is [\texttt{Answer}] \\
    \noalign{\hrule height 1.2pt}
    \end{tabular}
\caption{Prompt templates translating the original questions in ProtoQA to \textsc{NL} that fits the next-word prediction objective of LLMs.}
\label{tab:protoqa-templates}
\end{table*}

\end{document}